\documentclass[lettersize,journal]{IEEEtran}

\usepackage{acronym}
\usepackage{amsmath}
\usepackage{amssymb}
\usepackage{mathtools}
\usepackage{amsthm}
\usepackage{amsfonts}%
\usepackage{mathrsfs}%
\usepackage{hyperref}
\usepackage{url}
\usepackage[capitalize,noabbrev]{cleveref}
\usepackage{color, colortbl}
\usepackage{algorithm}
\usepackage{algorithmicx}%
\usepackage{algpseudocode}%
\usepackage{listings}%
\usepackage{graphicx}
\usepackage{booktabs}
\usepackage{paralist}
\usepackage[inline]{enumitem}

\usepackage{xcolor}%
\usepackage{textcomp}%
\usepackage{manyfoot}%
\usepackage{multirow}
\usepackage{caption}

\usepackage{array}
\usepackage[caption=false,font=normalsize,labelfont=sf,textfont=sf]{subfig}
\usepackage{stfloats}
\usepackage{verbatim}
\usepackage{cite}

\usepackage{bm}

\def\eqref#1{equation~\ref{#1}}

\def\1{\bm{1}}

\DeclareMathAlphabet{\mathsfit}{\encodingdefault}{\sfdefault}{m}{sl}
\SetMathAlphabet{\mathsfit}{bold}{\encodingdefault}{\sfdefault}{bx}{n}

\DeclareMathOperator*{\argmax}{arg\,max}

\newtheorem{proposition}{Proposition}

\acrodef{ODE}{Ordinary Differential Equation}
\acrodef{NODE}{Neural \acl{ODE}}
\acrodef{CDE}{Controlled Differential Equation}
\acrodef{NCDE}{Neural \acl{CDE}}
\acrodef{RNN}{Recurrent Neural Network}
\acrodef{LSTM}{Long Short-Term Memory}
\acrodef{XAI}{Explainable Artificial Intelligence}
\acrodef{KDE}{Kernel Density Estimation}
\acrodef{AMORE}{Automatic and Model-agnostic Regional Rule Extraction}
\acrodef{GNN}{Graph Neural Network}
\acrodef{CNN}{Convolutional Neural Network}
\acrodef{DT}{Decision Tree}

\begin{document}


\title{Enabling Regional Explainability by Automatic and Model-agnostic Rule Extraction}

\author{Yu Chen, Tianyu Cui, Alexander Capstick, Nan Fletcher-Loyd, Payam Barnaghi 
\thanks{All authors are with Department of Brain Sciences, Imperial College London, London, United Kingdom (e-mail: yu.chen@imperial.ac.uk; t.cui23@imperial.ac.uk; alexander.capstick19@imperial.ac.uk; nan.fletcher-lloyd17@imperial.ac.uk; p.barnaghi@imperial.ac.uk).}
\thanks{All authors are also with Care Research and Technology Centre, The UK Dementia Research Institute, London, United Kingdom.}
\thanks{Corresponding author: Payam Barnaghi.}
}







\maketitle

\begin{abstract}
In Explainable AI, rule extraction translates model knowledge into logical rules, such as IF-THEN statements, crucial for understanding patterns learned by black-box models. This could significantly aid in fields like disease diagnosis, disease progression estimation, or drug discovery. However, such application domains often contain imbalanced data, with the class of interest underrepresented. Existing methods inevitably compromise the performance of rules for the minor class to maximise the overall performance. As the first attempt in this field, we propose a model-agnostic approach for extracting rules from specific subgroups of data, featuring automatic rule generation for numerical features. This method enhances the regional explainability of machine learning models and offers wider applicability compared to existing methods. We additionally introduce a new method for selecting features to compose rules, reducing computational costs in high-dimensional spaces. Experiments across various datasets and models demonstrate the effectiveness of our methods.
\end{abstract}


\maketitle

\section{Introduction}

\ac{XAI} \cite{arrieta2020explainable} has garnered increasing attention in recent years, driven by the widespread adoption of deep learning models.
\ac{XAI} generally aims to enable professionals to understand why a model has made a specific prediction (decision) and/or what a model has learned from the data. For these purposes, prior works in \ac{XAI} often focus on producing either local or global \ac{XAI} \cite{li2020survey}. 
In global \ac{XAI}, Rule extraction (also known as rule learning) \cite{hailesilassie2016rule,he2020extract} is a field that translates a model's knowledge through a set of logical rules, such as IF-THEN statements, to obtain explainability of a model's reasoning. Such rule-based explanations can offer interpretations that mimic human experts' reasoning in solving knowledge-intensive problems \cite{grosan2011rule} and provide insights into key feature interactions in high dimensional spaces \cite{9448448}.
This type of interpretations are particularly desirable in knowledge-based applications of machine learning, such as drug discovery \cite{benet2016bddcs} and healthcare \cite{sutton2020overview}. 

\begin{figure*}
    \includegraphics[width=0.95\textwidth,trim=0.cm 0.cm 0.cm 0.cm,clip]{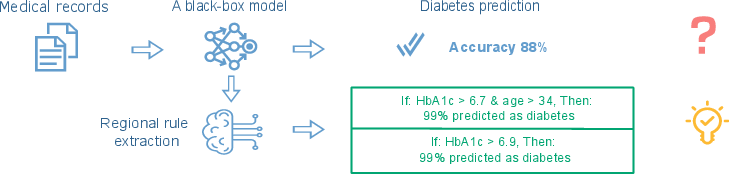}
    \caption{An example scenario of applying regional rule extraction for diabetes prediction. The data region of interest is the region that contains samples predicted as positive diabetes cases by the model. The results are from our experiments on a diabetes prediction task \cite{diabetes@kaggle}. In the extracted rules, Hemoglobin A1c (HbA1c) is a measure of a person's average blood sugar level over the past 2-3 months.}
    \label{fig:use_case}
\end{figure*}

In practice, we often want to interpret a model's knowledge for identifying a specified subgroup of the data. For instance, interpreting the patterns captured by a model for diagnosing a specific disease is highly desirable in clinical applications (\cref{fig:use_case}). Data samples labeled with the disease must share similar patterns that do not exist outside the specific subgroup. However, the subgroups of interest usually have sporadic occurrences in many real-world scenarios, as a minor class in classifications tasks. Most applications that prefer rule explanations, such as disease diagnosis, drug discovery, and financial fraud detection \cite{haixiang2017learning}, often face such a challenge. Existing methods in rule extraction focus on global explainability, which compromises the regional explainability of minor subgroups of data to achieve better overall performance across the entire data distribution.


\begin{figure*}
    \includegraphics[width=0.95\textwidth,trim=0.cm 0.cm 0.cm 0.cm,clip]{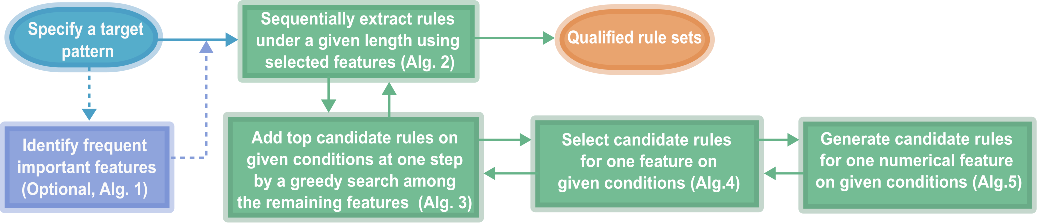}
    \caption{The workflow of our methods. \Cref{alg:find_freq_features} is optional, depending on the dimensionality of the data.}
    \label{fig:workflow}   
\end{figure*}


To better interpret data patterns for a specified subgroup, \emph{regional rule extraction} aims to extract optimal rules from a specified data region. Formally, the outcome of regional rule extraction from a model is in the form of IF-THEN statements: IF $x \in \mathbb{X}$, THEN $y \in \mathbb{Y}$, here $\mathbb{X}$ denotes a subspace of a multivariate variable $x$ confined by a set of rules, e.g. value ranges of several features. $\mathbb{Y}$ denotes a subset of a target variable $y$. For instance, when extracting rules from a classifier,  $y$ often refers to the model's prediction and $\mathbb{Y}$ denotes a class of interest (e.g. positive class in binary classification). 
 
Compared to global rule extraction, regional rule extraction does NOT generate rules for the data region $y \notin \mathbb{Y}$. 
It is expected to gain two advantages through this reduction: i) obtaining more accurate and generalized rules for the specified data region $y \in \mathbb{Y}$, and ii) providing more precise control on key properties of the extracted rules (e.g. number of rules, number of samples that satisfy these rules).

The general objective of regional rule extraction is to search for the optimal subspace of features that maximally elevates the occurrence of the data of interest. Conventional ways to extract rules from data often utilize various search strategies, such as hill-climbing, beam search, and exhaustive search \cite{molnar2023interpretable}.  Due to the vastness of the search space, these strategies often require predefined discretization for numerical features. However, predefined discretization is neither applicable nor desirable in many real-world applications. And in such a preprocessing, features are often discretised individually because the computational cost may increase exponentially when considering dependencies among a large number of features. Such individual discretization considers only the marginal distributions of features rather than their joint distributions, there is an implicit assumption that these features are independent -- a condition not valid in many real-world scenarios.

In order to enable regional explainability with easy access, we propose \emph{\ac{AMORE}} -- a novel approach that can automatically generate rules (e.g. key intervals of numerical features) through a model-agnostic algorithm. 
More specifically, our main contributions are three-fold: 
\begin{enumerate}[label=\roman*)]
\item we enhance regional rule extraction for black-box machine learning models, offering more accurate and generalized rules for underrepresented data regions than global rule extraction, and also allowing for local rule extraction for a given sample; 
\item we enable automatic and adaptive rule generation with numerical features by integrating the discretization process into the search strategy for a feature subspace, without requiring predefined discretization or making any assumption about feature distributions;
\item we develop an efficient method for selecting features to compose rules, which is designed to seamlessly integrate with our rule generation method and reduce computational cost in high-dimensional feature spaces.
\end{enumerate}
The workflow of our methods is illustrated in \Cref{fig:workflow}. To the best of our knowledge, this is the first attempt to specifically address regional rule extraction with automated rule generation in the literature of \ac{XAI}.

\section{Related Work}

Rule extraction is the most human-friendly solution in global \ac{XAI} for knowledge-intensive applications because it can directly provide human-understandable explanations as long as the data features are interpretable. This property is particular desirable for tabular data, which is prevalent in domains such as healthcare, finance, and social science. Nonetheless, most efforts in global \ac{XAI} have been focused on computer vision tasks \cite{zhang2018interpretable,yosinski2015understanding,achtibat2023attribution} and language processing tasks \cite{esser2020disentangling,shen2020interpreting,zhao2024explainability}, which primarily rely on visualization and semantic concepts to bring interpretability to deep models. To improve regional explainability for various types of data, we aim to develop a general method of rule extraction for identifying regional data patterns. 

There are three conceptual categories of the implementation of rule extraction in neural networks \cite{hailesilassie2016rule,he2020extract}: decompositional methods \cite{sato2001rule,wu2020regional,frosst2017distilling,lal2022nn2rules} -- the rule extraction is performed layer by layer and involves every neuron in a model; pedagogical methods \cite{saad2007neural,sethi2012kdruleex,de2015active} -- where only the input and output of a model are involved in the rule extraction; eclectic methods \cite{hruschka2006extracting,setiono2008recursive} -- the rule extraction can be performed through a combination of decompositional and pedagogical approaches, involving the grouping of components within a neural network. In this work, our focus is on pedagogical rule extraction which is model-agnostic.

One main challenge of rule extraction is the discretization of numerical features, with the goal of locating the most informative value intervals for numerical features. Most existing methods require pre-defined discretization to reduce the search space. For instance, association rule classifiers \cite{liu1998integrating,luo2016automatically} use association rule mining \cite{agrawal1994fast,soltani2014confabulation} to transform rule extraction into finding frequent item sets -- groups of items that frequently appear together in training samples. Consequently, numerical features must be transformed to discretized items before this process. Similarly, KDRuleEx \cite{sethi2012kdruleex} was developed to extract decision tables using categorical features. MUSE \cite{lakkaraju2019faithful} is a method for obtaining faithful rule sets in a feature subspace, where the candidate rules of numerical features must be readily provided. Moreover, Falling rule list \cite{wang2015falling}, Scalable Bayesian Rule Lists (SBRL) \cite{yang2017scalable}, and Bayesian Rule Sets (BRS) \cite{wang2017bayesian} are probabilistic rule extraction methods that use distributions of discrete variables for modeling rule generation. 

A few prior work made attempts to automate the discretization process. For example,  REFNE \cite{zhou2003extracting} utilizes ChiMerge \cite{kerber1992chimerge} to automatically discretize numerical features for extracting rules from model ensembles. It relies on an augmented data set that is generated by model ensembles to enable ChiMerge as an automated process for rule generation. In addition, it ignores interactions between features because each feature is discretized independently.

Among the existing methods with automated rule generation,  Decision Tree (DT) \cite{quinlan2014c4,de2000classification} is the most popular one applied in rule extraction for neural networks due to its flexibility and accessibility \cite{sato2001rule,wu2020regional,frosst2017distilling}. DTs are inherently pedagogical, yet they are often utilized in decompositional methods to construct decision paths between intermediate layers. Each decision path to a leaf node in a decision tree naturally represents a rule set associated with a data subgroup.  
However, DT is not well-suited for regional rule extraction because the criteria of splitting (segmenting) numerical features are based on a global objective that usually compromises performance on minor subgroups.  In our experiments, we found that up-weighting the minor class in DT can not improve it's performance on regional rule extraction. We posit this is because the selection of feature splits may still be affected by other data regions that are not of interest, especially in multi-mode scenarios. 

A relevant concept of regional rule extraction is sequential covering, which iteratively learns a rule list for a specific data subgroup until a decision set that covers the entire dataset is composed \cite{molnar2023interpretable}. In each step of sequential covering, learning the rule list for a specified subgroup is a typical scenario of regional rule extraction. 

\section{Methods}


In the rest of this article, we use following notations: $\mathbb{X}$ denotes a rule set with only conjunctive rules, $|\mathbb{X}|$ denotes the length of the rule set, $\mathbb{Y}$ denotes a specified value subset of the target variable $y$, $n$ is the sample index, $N$ is the number of training samples, and $\mathbb{I}(\cdot)$ denotes the indicator function.

Formally, the objective of regional rule extraction is to find a subspace of data features that maximizes the purity of a specified data subgroup within it. The feature subspace is defined by a rule set with two constraints: i) $l_{max}$ -- the maximum number of rules included in the rule set, constraining the complexity of the rule set to not be too large; ii) $s_{min}$ -- the minimum number of training samples that satisfy the rule set, constraining the support of the rules to not be too small. We quantify the purity of a specified data subgroup as its conditional probability within a subspace, then the objective can be formulated as:
\begin{equation}\label{eq:rr_obj}
    \begin{split}
        & \qquad \argmax_{\mathbb{X}} \Pr(y\in \mathbb{Y}|x \in \mathbb{X})  
        \\ 
        & s.t. \ \ |\mathbb{X}| \le l_{max}, \quad \sum_{n=1}^N \mathbb{I}(x^{(n)} \in \mathbb{X}) \ge s_{min} 
    \end{split}
\end{equation}
We define one rule by one feature, i.e. one value of a categorical feature or one value interval of a numerical feature. Thus, the length of a rule set is also the number of included features. 

It is computationally prohibitive to solve such an optimization problem exactly, especially for numerical features in high-dimensional space. 
The challenges are mainly due to: 
 \begin{enumerate*}[label=\roman*)]
 \item large search space for selecting feature combinations to compose rules;
 \item large search space for selecting value intervals of numerical features. 
 \end{enumerate*}
 We tackle these challenges heuristically through two procedures: i) a feature selection procedure, and ii) a rule extraction procedure that integrates rule generation and selection processes.


\begin{figure*}[!ht]
    \centering 
    \subfloat[Demonstration of selecting feature contribution threshold $\mathbf{j}_{th}$ that satisfy \cref{eq:score_thd} by stepwise scanning.]{\includegraphics[width=0.45\textwidth,trim=0.cm 0.cm 0.cm 0.cm,clip]{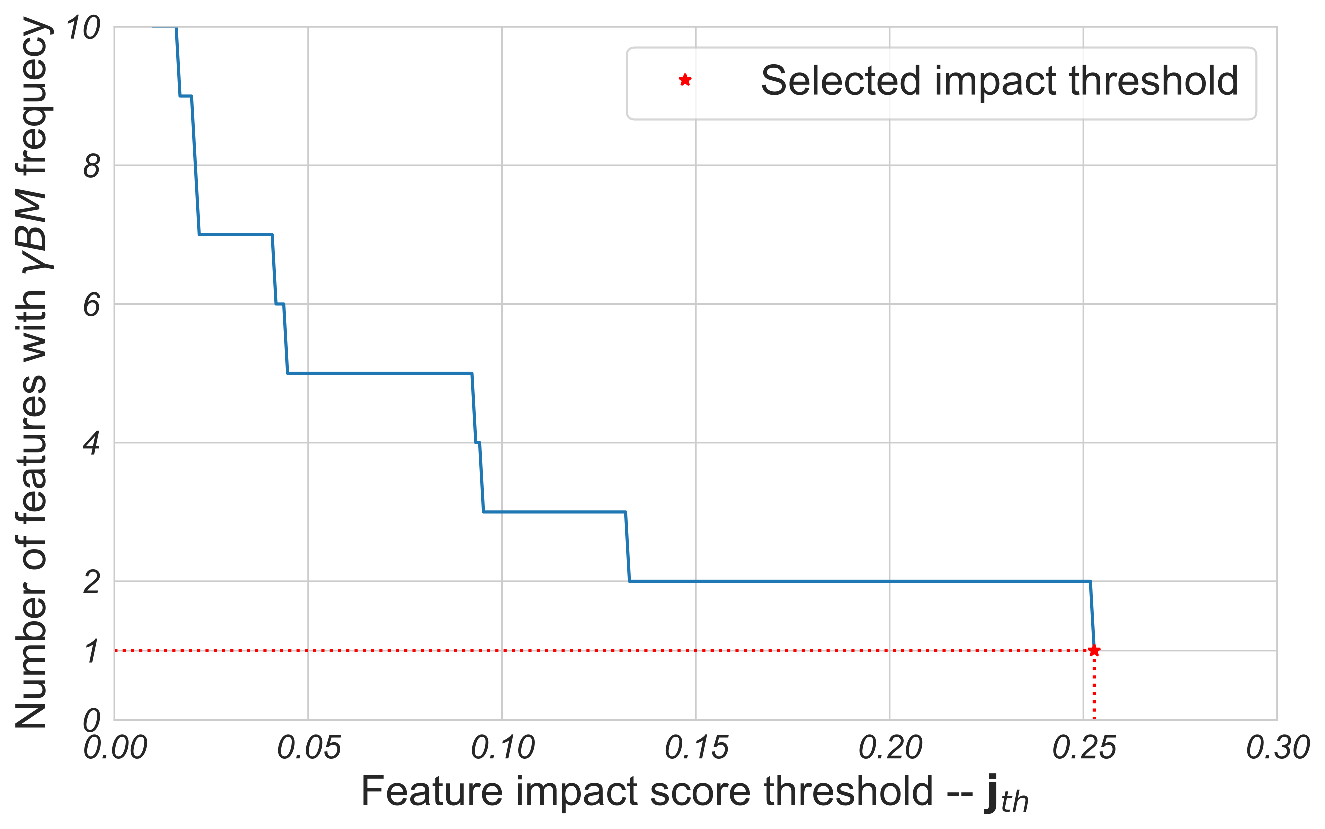}
    \label{fig:score_thd}}
    \hfill
    \subfloat[Illustration of generating feature intervals in a multi-mode scenario. We can capture both modes by initializing two intervals with the two peak grids.]{\includegraphics[width=0.45\textwidth,trim=0.cm 0.cm 0.cm 0 .5cm,clip]{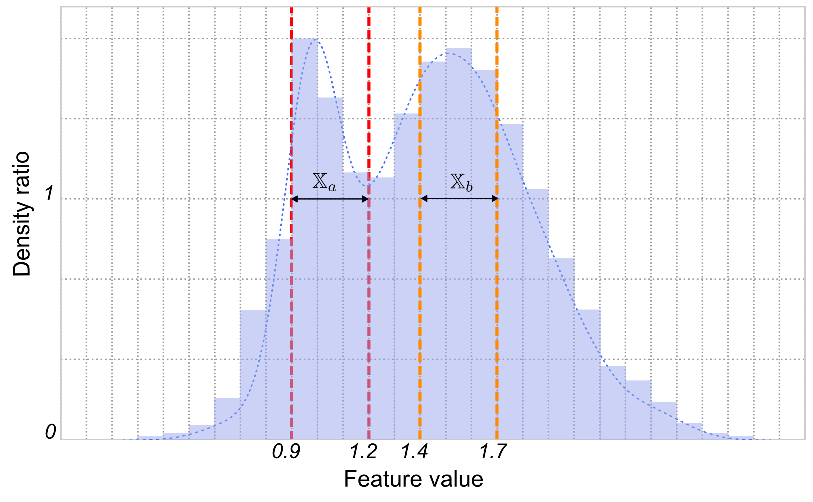}
    \label{fig:demo_feature_intervals}}
    \caption{Visualization of the threshold selection and feature interval searching in a multi-mode scenario.}
    \label{fig:methods}    
\end{figure*}

\subsection{Feature selection for rule extraction}

Although the feature selection procedure is optional, we recommend conducting it before rule extraction for high-dimensional data. 
To align with the purpose of rule extraction, we focus on selecting features that frequently appear in a decision path, aligning with the principles of decision-making. This can be achieved by mining feature sets that frequently have high importance across a sample set.

We propose a new method (\Cref{alg:find_freq_features}) for effectively and automatically selecting features of rules. Meanwhile, we introduce a new feature importance measurement (\Cref{eq:impact}) for differentiable models, leveraging integrated gradients \cite{sundararajan2017axiomatic} to quantify feature importance as it can better align with the principle of decision rules. For non-differentiable models, such as tree models or ensemble models, one can use any feature importance measures that are applicable, e.g., SHAP \cite{lundberg2017unified} or LIME \cite{ribeiro2016should}.

\subsubsection{Preliminary}

The integrated gradients (also known as integrated Jacobian) \cite{sundararajan2017axiomatic,crabbe2021explaining} is a pairwise measure that can quantify the contribution of changing an input feature on the shift of a target variable given by a trained differentiable model. It can be applied to identify feature sets that most frequently have a joint high impact on shifting a target variable. Such feature sets are crucial for determining the target variable's value. 

Assume we have a model $\mathcal{G}: \mathbb{R}^D \rightarrow \mathbb{R}$ to infer a variable $y \in \mathbb{R}$ by an observation $x \in \mathbb{R}^D$, i.e. $y = \mathcal{G}(x)$. Here $D$ is the number of features. 
Let $x_i$ denote the $i$-th feature of a baseline sample $x$, $\tilde{x}_i$ denote the $i$-th feature of a test sample $\tilde{x}$. We can compute the integrated gradient of the $i$-th feature with respect to the shift of $y$ between the two samples as below \cite{sundararajan2017axiomatic,crabbe2021explaining}:
\begin{equation}\label{eq:ijacob}
    \begin{split}
        \mathbf{j}_i &= (x_i - \tilde{x}_i)\int_{0}^{1} \frac{\partial \mathcal{G}(x)}{\partial x_i} \Bigr |_{x=\gamma_i(\lambda)} d\lambda, \\
        & \gamma_i(\lambda) = \tilde{x}_{i} + \lambda (x_i - \tilde{x}_i), \ \  \forall \lambda \in [0,1 ]
    \end{split}
\end{equation}
Here $\gamma_i(\lambda)$ represents a point between $x_i$ and $\tilde{x}_i$. 

\subsubsection{Feature importance measured by integrated gradients}
Based on \Cref{eq:ijacob}, we then estimate the importance of the $i$-th input feature on the shift of $y$ as the following equation:
\begin{equation}\label{eq:impact}
    \hat{\mathbf{j}}_{i} = \left|\frac{\mathbf{j}_{i}}{y-\tilde{y}}\right|, \quad \text{where} \ \ y = \mathcal{G}(x), \ \ \tilde{y} = \mathcal{G}(\tilde{x})
\end{equation}
In brief, $\hat{\mathbf{j}}_{i}$ represents the ratio of the shift attributed to the $i$-th feature out of the total shift caused by all features. 

As the original integrated gradients are sensitive to the scale of the target variable, we normalize the integrated gradients by the absolute value of the shift of the target variable. This normalization ensures that the importance of each feature is comparable across different sample pairs, which is crucial for identifying frequently important feature sets. 

Since this feature importance is measured using the shift between a baseline sample and a test sample, we suggest choosing class centers as baseline samples in classification tasks. 
Test samples should be a randomly selected subset from the training set to simulate decisions about the class origin of a random sample. We ensure the test samples are balanced across all classes, preventing the oversight of important features for minor subgroups of data . For regression or unsupervised tasks, the selection of baseline samples could depend on the specific application. A few general options are the mean, median, or mode of the training samples clustered by target subgroups.

\subsubsection{Frequently important feature sets}

The proposed feature importance (\Cref{eq:impact}) is measured by a baseline sample and a test sample. When given $B$ baseline samples, $M$ test samples, and $D$ features for each sample, one can produce an importance matrix $\mathbf{J} \in \mathbb{R}^{B \times M \times D}$. For instance, $\mathbf{J}_{b,m,i}$ is the importance of the $i$-th feature on the shift of the target variable from the $b$-th baseline sample to the $m$-th test sample. This importance matrix $\mathbf{J}$ can be readily transformed to a size of $B M \times D$. For feature importance determined by other measures, one may directly construct the importance matrix of size $N \times D$, where $N$ is the number of samples.

Then each row of the importance matrix $\mathbf{J}$ can be transformed into a sequence of feature indices, containing indices of features with an importance score greater than a threshold $\mathbf{j}_{th}$. Through this transformation, we are able to apply the FP-Growth algorithm \cite{han2000mining} to discover subsets of features that frequently have a high impact together. FP-Growth is an efficient algorithm for mining frequent item sets (i.e. items frequently appear in the same records). Here we treat a feature index as an item and one sequence as one record.


The impact threshold $\mathbf{j}_{th}$ is a key parameter for filtering less important features. If it is too small, too many less important features may be included. If it is too large, there may be not enough features that satisfy the minimum frequency requirement. In principle, we prefer a threshold that can provide enough features and be as high as possible. To determine the threshold automatically, we suggest selecting it by the following equation: 
\begin{equation}\label{eq:score_thd}
    \sum_{i=1}^D \mathbb{I}\left( \left(\sum_{b=1}^{B} \sum_{m=1}^{M} \mathbb{I}((\mathbf{J}_{b,m,i} \ge \mathbf{j}_{th})\right) \ge \gamma {BM}\right) = 1, 
\end{equation}
This equation requires the threshold $\mathbf{j}_{th}$ to be sufficiently large so that only one feature can have an impact score larger than $\mathbf{j}_{th}$ in most rows of the importance matrix $\mathbf{J}$. $\gamma BM$ specifies the minimum coverage of the most frequent feature, where $\gamma \in (c_{min}/BM,1]$, and $c_{min}$ is the minimum occurrence of a frequent set. For instance, when $\gamma=1$, it requires that the most frequent feature has impact scores larger than $\mathbf{j}_{th}$ for all samples, similar to the root node of a decision tree. We set $\gamma = 0.99$ in all of our experiments. When $\gamma$ is smaller, $\mathbf{j}_{th}$ will be higher, and more features will be filtered out. 
Our implementation uses a stepwise scanning process to find the threshold. As illustrated in \Cref{fig:score_thd}, we initialize the threshold with a small value and increase it stepwise until only one feature has a frequency no less than $\gamma BM$ (\cref{eq:score_thd}). It is clear to see that the number of features that satisfy the $\gamma BM$ frequency requirement decreases as the threshold increases. In this way, an appropriate threshold can be automatically selected to include sufficiently representative features for rule extraction. The pseudo-code of feature selection is described in \cref{alg:find_freq_features}.
 
In FP-Growth algorithm, a frequent item set should have appeared at least $c_{min}$ times among all records. Additionally, it is often required to include fewer than $k_{max}$ items. This requires $c_{min}$ and $k_{max}$ as two hyperparameters for the FP-Growth algorithm. $c_{min}$ determines the least frequent item in a frequent item set, so we suggest a relatively small value for it, such as 10\% of the training size. The use of threshold $\mathbf{j}_{th}$ automatically limits the number of items in a frequent item set, so we just set $k_{max}$ to be large enough to encompass the longest item set.  When there are multiple frequent item sets discovered, we choose the longest one, and if there are ties, we choose the most frequent one as the candidate feature set for rule extraction.

\begin{algorithm}
    \caption{FrequentFeaturesSelection}
    \label{alg:find_freq_features}
    \begin{algorithmic}
    \small
    \State {\bfseries Input}: $X_b$ -- baseline samples, $X_t$ -- test samples, $\mathcal{G}$ -- a model producing the target variable $y$, $c_{min}$ -- minimum frequency of a frequently important feature set, $k_{max}$ -- maximum length of a frequently important feature set.
    \State
    \State $\mathbf{J} \leftarrow$  generate the importance matrix by $X_b$, $X_t$, $\mathcal{G}$ (\Cref{eq:ijacob,eq:impact}),
    \State $\mathbf{j}_{th} \leftarrow$ decide the threshold of impact score (\Cref{eq:score_thd}),
    \State $\mathbf{F} \leftarrow$ transform $\mathbf{J}$ to a set of feature sequences, each sequence contains the indices of features having an impact score larger than $\mathbf{j}_{th}$;   
    \State $F_{fq} \leftarrow$ obtain frequently important feature sets by FP-Growth($\mathbf{F},c_{min},k_{max}$) \cite{han2000mining}
    \State $F^{*}_{fq} \leftarrow$ the longest feature set in $F_{fq}$, if there are ties, choose the most frequent one;
    \State {\bfseries Return:} $F^{*}_{fq}$
    \end{algorithmic}
\end{algorithm}

\subsection{\acf{AMORE}}

\begin{figure*}
    \centering
    \includegraphics[width=0.75\textwidth,trim=0.cm 0.cm 0.cm 0.cm,clip]{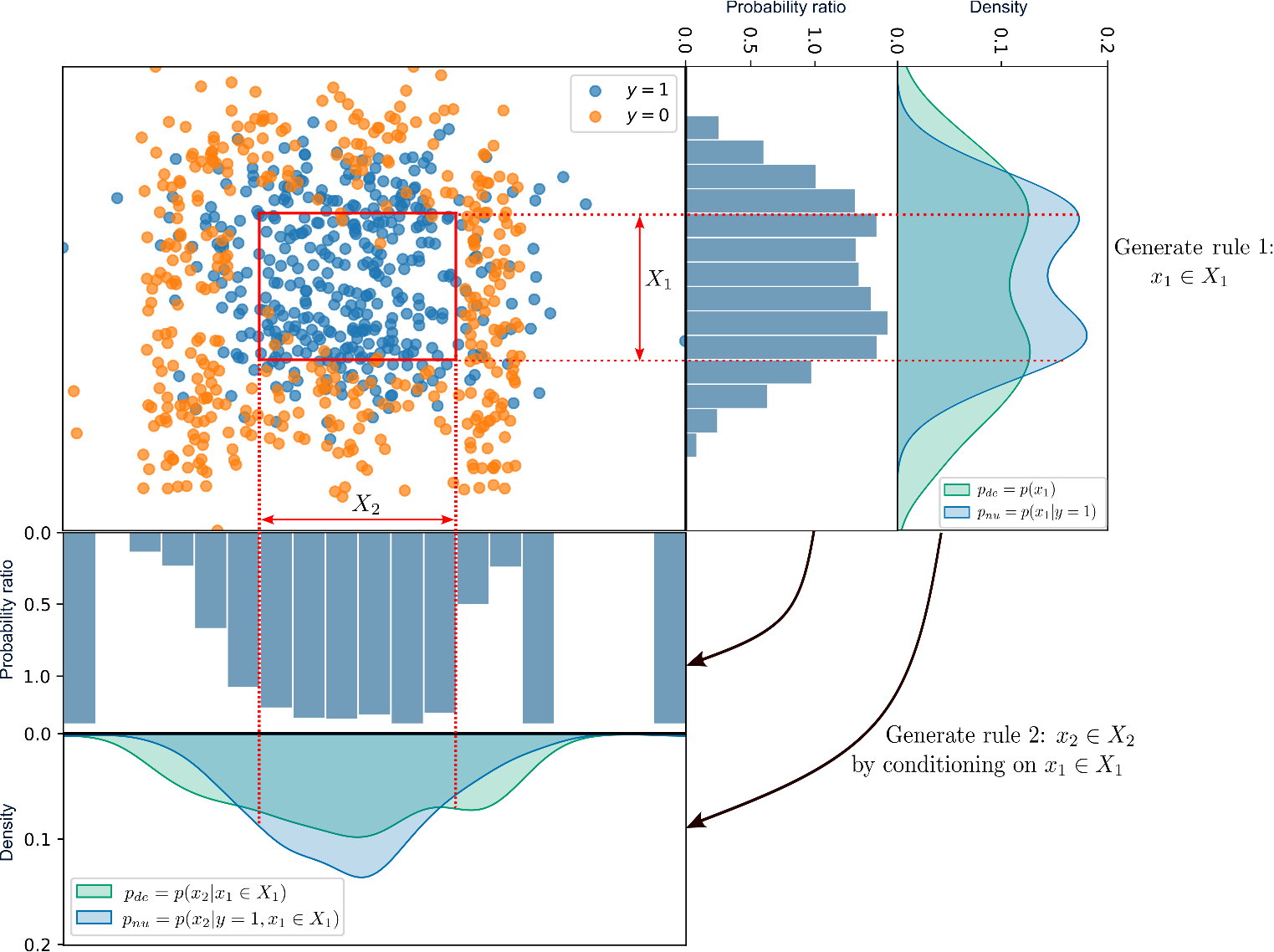}
    \caption{Demonstration of the automatic rule generation with numerical features. This figure was plotted with synthetic data of two classes in a 2-dimensional feature space. The target subgroup is set as $y=1$. We generate the first rule by identifying the mode interval of the probability ratio on the feature $x_1$ (the vertical axis). The second rule for the feature $x_2$  (horizontal axis) is generated by conditioning on the first rule. We applied \ac{KDE} for plotting density curves, the histograms of probability ratios and the subspace in the red rectangle were obtained through our method. More details can be found in \Cref{alg:add_potential_rules,alg:raise_rule_interval}.}
    \label{fig:scatter_density_mode}
\end{figure*}

In general, regional rule extraction is to provide a subset of features with a length $\le l_{max}$ and the value ranges of those features that can maximize the conditional probability in \Cref{eq:rr_obj}. We can explicitly decompose the rule set to each selected feature, i.e. $\{x \in \mathbb{X}\} \triangleq \{x_{f_1} \in \mathbb{X}_{f_1}, \dots, x_{f_l} \in \mathbb{X}_{f_l}\}$, where $\{f_1,\dots,f_l\}$ are the indices of selected features, $\mathbb{X}_{f_i}$ $(\forall 1 \le i \le l)$ denotes a value range of $f_i$-th feature. According to the chain rule of conditional probability, we can then approximate the conditional probability as below: 
\begin{equation}\label{eq:ratio_prod}
    \begin{split}
        & \Pr(y \in \mathbb{Y}|x_{f_1} \in \mathbb{X}_{f_1}, \dots, x_{f_l} \in \mathbb{X}_{f_l}) \\
        & \propto \prod_{i=1}^{l} \frac{ \Pr(x_{f_i}\in \mathbb{X}_{f_i}|y\in \mathbb{Y}, x \in  \mathcal{X}_{l}^{i-1})}{\Pr(x_{f_i}\in \mathbb{X}_{f_i}| x \in {\mathcal{X}}_{l}^{i-1})}\\
        &   \{x \in \mathcal{X}_{l}^{i-1}\} \triangleq \{x_{f_1} \in \mathbb{X}_{f_1},\dots, x_{f_{i-1}} \in \mathbb{X}_{f_{i-1}}\}, \\ 
        & \quad \forall 1 < i \le l, \quad  \mathcal{X}_{l}^{0} \triangleq \text{dom}(x)       
    \end{split}
\end{equation}
$\mathcal{X}_{l}^{i-1}$ denotes conditions composed by previous $i-1$ features and $\mathcal{X}_{l}^{0}$ is the domain of $x$. 
We treat this decomposition as a sequence of operations. In the $i$-th operation, we select the value range of $f_i$-th feature that maximizes a probability ratio $\mathbf{r}_{f_i}$, where $\mathbf{r}_{f_i}$ is defined as the following:
\begin{equation}
    \begin{split}
        & \mathbf{r}_{f_i} \triangleq  \frac{\Pr(x_{f_i}\in \mathbb{X}_{f_i}|y\in \mathbb{Y}, x \in  \mathcal{X}_{l}^{i-1})}{\Pr(x_{f_i}\in \mathbb{X}_{f_i}| x \in \mathcal{X}_{l}^{i-1})}  
    \end{split}
    \label{eq:prob_ratio}
\end{equation}
Through this decomposed objective for each feature, we can obtain candidate rules one by one for a given order of features. However, the result is not guaranteed to be globally optimal if we do not search all possible orders of features. As it is usually computationally prohibitive to find the exact global optimal solution, a feasible way is to select the feature $f_i$ that has the maximum $\mathbf{r}_{f_i}$ among all remaining features at the $i$-th step, which is analogous to the strategy of adding branches in decision trees. This approach can achieve a global optimal solution when the features are independent of each other, as the order of features does not matter in such cases.

Without any assumptions about the conditional distributions, we estimate the ratio $\mathbf{r}_{f_i}$ using training samples as below:
\begin{equation}\label{eq:ratio_approx}
\begin{split}
    & \mathbf{r}_{f_i}  \approx  \frac{\sum_{n=1}^{N} \mathbb{I}(x^{(n)} \in \mathcal{X}_{l}^{i-1})}{\sum_{n=1}^{N} \mathbb{I}(x^{(n)}_{f_i}\in \mathbb{X}_{f_i} \ \ \& \ \ x^{(n)} \in \mathcal{X}_{l}^{i-1})} \times
    \\ &  
    \frac{\sum_{n=1}^{N} \mathbb{I}(x^{(n)}_{f_i} \in \mathbb{X}_{f_i} \ \ \& \ \ (x^{(n)},y^{(n)}) \in (\mathcal{X}_{l}^{i-1},\mathbb{Y}) )}{\sum_{n=1}^{N} \mathbb{I}((x^{(n)},y^{(n)}) \in (\mathcal{X}_{l}^{i-1},\mathbb{Y}))},\\
\end{split}
\end{equation}
One advantage of using this probability ratio as our stepwise objective is that we can directly discard candidate rules with a ratio not larger than 1 because it does not increase the overall conditional probability in \cref{eq:ratio_prod}.

We provide detailed pseudo code of this process in \cref{alg:extract_rule_sets,alg:add_rule_branch}. The input argument $n_g$ is the number of grids for initializing value intervals of numerical features, which will be explained in \cref{alg:add_potential_rules}. $K$ is the maximum number of candidate rules that can be added at each step. We may obtain multiple rule sets when $K > 1$ to cope with multi-mode scenarios in data distributions. Note that there may be fewer than $K$ rules added at each step as we discard rules with a ratio not larger than 1. We set $K = 3$ in all of our experiments.

\begin{algorithm}
    \caption{ExtractRuleSets}
    \label{alg:extract_rule_sets}
    \begin{algorithmic}
    \small
    \State {\bfseries Input: 
    $X$ -- features of training samples;
    ${Y}$ -- indicator of the target subgroup $y \in \mathbb{Y}$, which is a binary vector; 
    $F^{*}_{fq}$ -- the selected feature set;
    $s_{min}$ -- minimum support of a rule set; 
    $l_{max}$ -- maximum length of a rule set;
    $n_{g}$ -- number of value grids for numerical features;
    $K$ -- maximum number of candidate rules added at each step.} 
    
    \State
    \State $\mathcal{R}_{tree} \leftarrow$ Initialise a tree structure for storing candidate rule sets,
    \State $\mathcal{X}^{0}_{l_{max}} \leftarrow$ Initialise previous rules by domain of $x$, 
    \State \# \emph{Start adding rules from the root node, see \cref{alg:add_rule_branch}.}
    \State {AddRules}($\mathcal{R}_{tree}.root$, $F^{*}_{fq}$, $X$, $\mathbb{Y}$, $\mathcal{X}^{0}_{l_{max}}$, $s_{min}$, $n_g$, $l_{max}$, $K$),
    \State $\mathbb{R}_{sets} \leftarrow $ Obtain all candidate rule sets in $\mathcal{R}_{tree}$
      
    \State {\bfseries Return:} $\mathbb{R}_{sets}$
    \end{algorithmic}
\end{algorithm}

\begin{algorithm}
    \caption{AddRules}
    \label{alg:add_rule_branch}
    \begin{algorithmic}
    \small
    \State {\bfseries Input: 
    $\bm{\nu}_{par}$ -- parent rule node; 
    $F^{*}_{fq}$ -- the selected feature set;
    $X$ -- features of training samples;
    ${Y}$ -- indicator of the target subgroup $y \in \mathbb{Y}$, which is a binary vector; 
    $\mathcal{X}_{pre}$ -- previous rules; 
    $s_{min}$ -- minimum support of a rule set; 
    $l_{max}$ -- maximum length of a rule set;
    $n_{g}$ -- number of value grids for numerical features;
    $K$ -- maximum number of candidate rules added at each step.} 
    
    \State
    \State $\tilde{F}^{*}_{fq} \leftarrow$ Remove the feature of the parent rule node $\bm{\nu}_{par}$ in $F^{*}_{fq}$,
    \If{$\tilde{F}^{*}_{fq}$ is not empty}
        \State $\mathbb{X}^{*}_K \leftarrow$ {obtain candidate rules for each feature in $\tilde{F}^{*}_{fq}$, and select the top $K$ rules with highest ratios, see \cref{alg:add_potential_rules} for getting candidate rules for one numerical feature.}      
        \For{$\mathbb{X}^{*}_{f_i}\in \mathbb{X}^{*}_K$}
            \State $\bm{\nu}_{new} \leftarrow$ Create a new rule node by $\mathbb{X}^{*}_{f_i}$, 
            \State $\tilde{\mathcal{X}}_{pre} \leftarrow$ Update ${\mathcal{X}}_{pre}$ by adding $\mathbb{X}^{*}_{f_i}$.
            \If{$len(\tilde{\mathcal{X}}_{pre}) < l_{max}$}
            \State AddRules($\bm{\nu}_{new}$, $\tilde{F}^{*}_{fq}$, $X$, $\mathbb{Y}$, $\tilde{\mathcal{X}}_{pre}$, $s_{min}$, $n_g$, $l_{max}$, $K$)
            \EndIf
            
        \EndFor
            
    \EndIf
    \end{algorithmic}
\end{algorithm}

\subsubsection{Generating rules for numerical features}

To generate candidate rules for categorical features, each category can be considered as a candidate for the designated rule. However, it is not trivial to generate candidate rules for numerical features. 
The question here is how to automatically identify intervals of numerical features that have high enough probability ratios and large enough support. We propose a histogram-based approach that initiates a value interval by a specified grid and iteratively expands the interval to its neighbor grids when certain criteria are met. 

This process begins with a binning step that divides the value range of a feature into a number of grids $n_g$. We offer a few options for the binning strategy in our implementation, including ``uniform",``kmeans", and ``quantile". The ``uniform" strategy ensures all initial grids in each feature have identical widths; ``kmeans" assigns values in each grid have the same nearest centre of a 1D k-means cluster; ``quantile" results in grids with an equal number of points across each feature. We found that the ``uniform" and ``kmeans" strategies work better than ``quantile" in all of our experiments.

Intuitively, we should start with a peak grid that gives the highest ratio $\mathbf{r}_{f_i}$ and expand it when possible. Nonetheless, this might not be able to find the optimal interval if the ratios in its neighbor grids decrease rapidly. An example is demonstrated in \Cref{fig:demo_feature_intervals}, in which the peak grid in the interval $\mathbb{X}_{a}$ is higher than the peak grid in the interval $\mathbb{X}_{b}$, however, the overall ratio of the interval $\mathbb{X}_{a}$ is lower than the overall ratio of the interval $\mathbb{X}_{b}$. It is also possible that there exist multiple intervals with similar ratios and we can not know which one is better before finishing the search for a whole rule set.
Considering such multi-mode scenarios, we search intervals that start with every peak grid and choose candidate intervals with the top $K$ ratios. We implemented a simple method based on first-order derivatives to find peaks of the ratio histogram, which can be replaced by more advanced methods. We only select peak grids with a ratio larger than 1. 
Here $K$ is the same as the maximum number of candidate rules can be added in each step (\Cref{alg:add_rule_branch}).  

\begin{algorithm}
    \caption{GetCandidateRules}
    \label{alg:add_potential_rules}
    \begin{algorithmic}
    \small
    \State {\bfseries Input: $X$ -- features of training samples;
    $f$ -- index of the specific feature;
    ${Y}$ -- indicator of the target subgroup $y \in \mathbb{Y}$, which is a binary vector; 
    $\mathcal{X}_{pre}$ -- previous rules;
    $s_{min}$ -- minimum support of a rule set; 
    $n_{g}$ -- number of value grids for numerical features;
    $K$ -- maximum number of candidate rules.} 
    
    \State
    \State $\mathbf{x}^{f}_{g} \leftarrow$ get $ n_{g}$  value grids of $f$-th feature 
    \State $\mathbf{r}_{g} \leftarrow$ Calculate probability ratios of each grids (\Cref{eq:ratio_approx}),
    \State $\mathbf{s}_{g} \leftarrow$ Calculate supports of each grids, 
    \State $\mathbf{x}^{f}_{g},\mathbf{r}_{g},\mathbf{s}_{g} \leftarrow$ Merge consecutive grids with identical ratios, merge empty grids to its neighbour grid with a higher ratio, 
    \State $\mathbf{g}_{p} \leftarrow $ Get peak grids with ratio peaks larger than 1 and sort them by their ratios, 

    \State $\mathbb{X}_f, \mathbf{r}_f \leftarrow$ {Generate candidate rules for $f$-th feature by \cref{alg:raise_rule_interval}, starting with every grids in $\mathbf{g}_{p}$.}    
    \State {\bfseries Return:} $K$ rules in $\mathbb{X}_{f}$ with top $K$ ratios in $\mathbf{r}_{f}$
    \end{algorithmic}
\end{algorithm}

After initiating an interval by specifying a peak grid, we then expand it by the following criteria: 
\begin{enumerate}[label=\roman*)]
    \item when the support of this interval is smaller than the minimum support $s_{min}$ we expand the interval to its neighbor grids if possible;
    \item when the support of this interval is not smaller than $s_{min}$, we only expand the interval to a neighbor grid when this neighbor grid has a higher ratio. This will guarantee that an interval with enough support only expands when its ratio gets increased (\Cref{prop1}).
\end{enumerate}  
The pseudo-code of this rule generation process for numerical features are described in \cref{alg:add_potential_rules,alg:raise_rule_interval}. \Cref{fig:scatter_density_mode} visualizes this process with synthetic data in a 2-dimensional feature space. 

\begin{proposition}\label{prop1}
    Given two exclusive intervals of a feature: $\mathbb{X}_{a}$, $\mathbb{X}_{b}$, where $\mathbb{X}_{a} \cap \mathbb{X}_{b} = \emptyset$, and their corresponding ratios $\mathbf{r}_{a}, \mathbf{r}_{b}$, if $0 \le \mathbf{r}_{a} < \mathbf{r}_{b}$, then the merged interval $\mathbb{X}_{m} = \mathbb{X}_{a} \cup \mathbb{X}_{b}$ has a ratio $\mathbf{r}_{m}$ that satisfies $\mathbf{r}_{a} < \mathbf{r}_{m} < \mathbf{r}_{b}$. 
\end{proposition}
\begin{proof}
    We first rewrite the density ratio as $\mathbf{r}_{a} = {\alpha_a}/{\beta_a}$, $\mathbf{r}_{b} = {\alpha_b}/{\beta_b}$, where $\alpha_a, \alpha_b$ are non-negative integers, $\beta_a,\beta_b$ are positive integers. According to \Cref{eq:prob_ratio}, then we have $\mathbf{r}_{m} = (\alpha_a+\alpha_b)/(\beta_a+\beta_b)$. So we can have:
   \begin{equation*}
       \begin{split}
       \mathbf{r}_{m} -  \mathbf{r}_{a} &=  \frac{\alpha_a+\alpha_b}{\beta_a+\beta_b} - \frac{\alpha_a}{\beta_a} \\
       &= \frac{\alpha_b - \alpha_a \frac{\beta_b}{\beta_a}}{\beta_a+\beta_b} 
       = \frac{\frac{\alpha_b}{\beta_b}-\frac{\alpha_a}{\beta_a}}{\frac{\beta_a}{\beta_b}+1} > 0
       \end{split}
   \end{equation*}   
Similarly, we can prove $\mathbf{r}_{m} -  \mathbf{r}_{b} < 0$, which completes the proof.
\end{proof}

\begin{algorithm}
    \caption{GenFeatureInterval}
    \label{alg:raise_rule_interval}
    \begin{algorithmic}
    \small
    \State {\bfseries Input: $s_{min}$ -- minimum support of a rule set,
    $g_p$ -- index of a peak grid to initiate an interval;
    $\mathbf{x}_{g}$ -- value grids of a feature;  
    $\mathbf{r}_{g}$ -- probability ratios of value grids; 
    $\mathbf{s}_{g}$ -- supports of value grids;
    ${Y}$ -- indicator of the target subgroup $y \in \mathbb{Y}$, which is a binary vector;
    $\mathcal{X}_{pre}$ -- previous rules.}
    
    \State
    \State $\mathbb{X}_{f,p}, s, r \leftarrow$ Initialize an interval by the peak grid $g_p$ and its support $\mathbf{s}_{g}[g_p]$, ratio $\mathbf{r}_{g}[g_p]$,
    \While{$s < s_{min}$ or $r > 1$}
        \If{$s < s_{min}$}
        \State $\mathbb{X}_{f,p}, s, r \leftarrow$ Expand the interval to a neighbor grid that has a larger ratio than the other neighbor grid,
        \ElsIf{$r > 1$}
        \State $\mathbb{X}_{f,p}, s, r \leftarrow$ Expand the interval to a neighbor grid that has a higher ratio than the other neighbor grid and also higher than $r$. If no such grid exists, break.
        \EndIf
    \EndWhile

    \If{$s < s_{min}$ or $r \le 1$}        
    \State $\mathbb{X}_{f,p} = \emptyset$
    \EndIf  

    \State {\bfseries Return:} $\mathbb{X}_{f,p},r, s$
    \end{algorithmic}
\end{algorithm}

Additionally, we provide local rule extraction for a given sample (as demonstrated in \cref{tab:diabetes_local} in Results section), which forces the value interval of a feature to include the feature value of a given sample. If the feature intervals obtained by \cref{alg:add_potential_rules,alg:raise_rule_interval} do not meet this requirement, we apply a local searching process. This can be directly achieved by starting with the grid that includes the given sample. The local rule extraction may not be able to obtain any valid rules (i.e. the ratios of any value intervals are not greater than 1) if the given sample is an outlier of the target subgroup.

\subsection{Evaluation criteria}

We apply three criteria for evaluating the quality of an extracted rule set $\mathbb{X}$ from a model. As we conduct experiments on classification tasks, we denote the true label of a sample as $y^*$, the model's prediction as $\hat{y}$, the class of interest as the target class $c$, the target subgroup for rule extraction is given by the model's prediction as $\hat{y} = c$.
\begin{enumerate}
    \item \emph{Support} -- the number of training samples that satisfy the rule set. It is a metric for measuring the coverage of the rule set, acting a role like recall in classification evaluation. A larger support is not necessarily better as there is usually a trade-off between support and confidence.
    \begin{equation*}
        \text{Support} = \sum_{n=1}^N\mathbb{I}(x^{(n)} \in \mathbb{X})
    \end{equation*}
    \item \emph{Confidence} (also called rule accuracy) \cite{luo2016automatically} -- the proportion of samples that belong to the target subgroup while satisfying the rule set. It measures the purity of the target subgroup given the rule set: 
    \begin{equation*}
        \text{Confidence} = \frac{\sum_{n=1}^N\mathbb{I}((\hat{y}^{(n)}=c) \& (x^{(n)} \in \mathbb{X}))}{\sum_{n=1}^N\mathbb{I}(x^{(n)} \in \mathbb{X})}
    \end{equation*}
    \item \emph{Fitness} -- a new metric we propose for regional rule extraction, being the ratio difference between samples that belong to the target subgroup and those that do not, under the rule set, divided by the total number of samples in the target subgroup. The higher the score, the better, and in the best case $\text{Fitness} = 1$:
    \begin{equation*}
    \begin{split}
        \text{Fitness} &= \frac{\sum_{n=1}^{N}\mathbb{I}((x^{(n)} \in \mathbb{X}) \& (\hat{y}^{(n)}=c))}{\sum_{n=1}^N\mathbb{I}(\hat{y}^{(n)} = c)} 
        \\
        & \quad - \frac{\sum_{n=1}^{N}\mathbb{I}((x^{(n)} \in \mathbb{X}) \& (\hat{y}^{(n)}\neq c))}{\sum_{n=1}^N\mathbb{I}(\hat{y}^{(n)} = c)}   \end{split}
    \end{equation*}    

\end{enumerate}
Support and confidence are widely applied in the literature of rule extraction. Typically, support and confidence are negatively correlated because including more samples likely introduces additional noise into the subspace. A better rule set should exhibit reduced correlation by increasing confidence while maintaining a similar support.
We define \emph{fitness} for evaluating regional rule sets, which takes into account the trade-off between support and confidence. This metric measures how well the subspace defined by a rule set fits into the target data region. A rule set with larger support and lower confidence, or one with smaller support and higher confidence, could both result in a lower fitness. Fitness can be negative when the subspace is mostly out of the target subgroup.
\section{Results}

\setlength{\tabcolsep}{2pt}
\renewcommand{\arraystretch}{1.2}
\setlength{\aboverulesep}{0pt}
\setlength{\belowrulesep}{0pt}

We conducted experiments on several tasks with a set of publicly available datasets and various models to demonstrate the effectiveness of the proposed methods, including: 
\begin{enumerate}[label= \roman*)]
\item diabetes prediction \cite{diabetes@kaggle} using a \emph{logistic regression model}; 
\item sepsis prediction \cite{reyna2019early} using a \emph{\ac{NCDE} model} \cite{kidger2020neural}; 
\item molecular toxicity prediction \cite{wu2018moleculenet} using a \emph{\ac{GNN}};
\item handwriting digits (MNIST \cite{lecun2010mnist}) classification using a \emph{\ac{CNN}};
\item brain tumour MRI image classification \cite{chitnis2022brain} using an \emph{EfficientNet} \cite{tan2019efficientnet} pre-trained on ImageNet-1K dataset \cite{imagenet15russakovsky}.
\end{enumerate}
A summarised description of all tasks is provided in Table 1 in Supplementary material.

Most of these tasks are imbalanced binary classification, a common scenario in applications requiring rule explanation. As a result, we observe that models often exhibit significantly better recall than precision even with techniques of up-weighting the minority class. 
In all these binary classification tasks, the positive class (i.e. $c = 1$) is designated as the class of interest. For obtaining a more balanced performance, we choose the prediction threshold $y_{th}$ by maximizing the difference between the True Positive Rate (TPR) and False Positive Rate (FPR) (i.e., $\text{TPR} - \text{FPR}$) using the ROC curve on the training set. Consequently, we consider the model predicting a positive sample when the model's output is greater than the threshold: $\hat{y}_p > y_{th}$, where $\hat{y}_p$ represents the probability of a sample being from positive class given by the model.
The MNIST and brain tumour MRI dataset are used to illustrate how our method can be applied to multi-classification tasks and how it can interpret image data through latent states. For both tasks, we select the class with the maximal $\hat{y}_{p}$ as the predicted class.

\begin{table*}[!ht]
    \centering
    \caption{Rules extracted from different datasets and models by our method \ac{AMORE} and Decision Tree (DT) classifier. In general, \ac{AMORE} gives much higher confidence criteria than DT classifier while showing higher fitness even with less support than DT. The results suggest that our method gains a better trade-off between support and confidence for a regional explanation.}
    \footnotesize
    \begin{tabular}{|c|c|c|c|}
    \toprule
    \rowcolor[HTML]{D6D6D6}
    \textbf{$\bm{l_{max}}$}  & \textbf{Rule properties} & \textbf{AMORE} (Ours) & \textbf{DT Classifier} \\
    \toprule
    \rowcolor[HTML]{E6E6E6}
    \multicolumn{4}{|c|}{\textbf{Diabetes prediction}}
    \\
    \midrule
    \multirow{4}{*}{1} & 
    \begin{tabular}[c]{@{}c@{}} Rules ($\mathbb{X}$) \\
    Target: $c=1$ \end{tabular} & 
    \begin{tabular}[l]{@{}l@{}} \texttt{HbA1c\_level} $ \ge 6.643$  \end{tabular} &  
    \begin{tabular}[l]{@{}l@{}} \texttt{HbA1c\_level} $ > 6.349$,\\
    \end{tabular} \\
    \cmidrule{2-4}
    & Support & 2736 & 14501    \\
    \cmidrule{2-4}
    & Confidence & 0.993 & 0.515  \\
    \cmidrule{2-4}
    & Fitness & 0.202 & 0.033    \\
    \midrule
    \rowcolor[HTML]{E6E6E6}
    \multicolumn{4}{|c|}{\textbf{Sepsis prediction}}
    \\
    \midrule
    \multirow{4}{*}{2} & \begin{tabular}[c]{@{}c@{}} Rules ($\mathbb{X}$) \\
        Target: $c = 1$
         \end{tabular} & 
        \begin{tabular}[c]{@{}c@{}} \texttt{HR\_ctime\_t45} $\ge 34$, \\
         \texttt{Temp\_ctime\_t38} $ \ge 34$
        \end{tabular} &  
        \begin{tabular}[c]{@{}c@{}} 
        \texttt{SBP\_ctime\_t22} $ > 9.5$,  \\
        \texttt{Temp\_ctime\_t21} $> 18.5 $
         \end{tabular} \\
        \cmidrule{2-4}
        & Support & 2068  & 2000    \\
        \cmidrule{2-4}
        & Confidence & 0.860   & 0.846  \\
        \cmidrule{2-4}
        & Fitness & 0.205 & 0.19    \\
    \midrule
    \rowcolor[HTML]{E6E6E6}
    \multicolumn{4}{|c|}{\textbf{Molecular toxicity prediction -- AR}}
    \\ \midrule
    \multirow{4}{*}{2} &
    \begin{tabular}[c]{@{}c@{}} Rules ($\mathbb{X}$) \\
    Target: $c = 1$ \end{tabular}    & \multicolumn{1}{c|}{\begin{tabular}[c]{@{}c@{}}$1.149 \le$ \texttt{NumAliphaticRings} $\le 10.405$, \\
     \texttt{SMR\_VSA5} $\ge 0.167$ \end{tabular}}     &        \multicolumn{1}{c|}{\begin{tabular}[c]{@{}c@{}} \texttt{NumAliphaticRings} $> 1.414$, \\
     \texttt{SMR\_VSA5} $> 0.836$ 
    \end{tabular}}                  \\ 
    \cmidrule{2-4}
    & Support & \multicolumn{1}{c|}{326}     &   253                          \\
    \cmidrule{2-4}
    & Confidence        & \multicolumn{1}{c|}{0.801}    &        \multicolumn{1}{c|}{0.874}               \\ 
    \cmidrule{2-4}
    & Fitness & 0.235 & 0.227    \\
    \midrule
    \rowcolor[HTML]{E6E6E6}
    \multicolumn{4}{|c|}{\textbf{Molecular toxicity prediction -- AhR}} \\
    \midrule
    \multirow{4}{*}{2} &
    \begin{tabular}[c]{@{}c@{}} Rules ($\mathbb{X}$) \\
    Target: $c = 1$ \end{tabular}    & \multicolumn{1}{c|}{\begin{tabular}[c]{@{}c@{}} $ 1.631 \le $\texttt{NumAliphaticCarbocycles}$\le 4.679 $, \\
     $1.341 \le $\texttt{fr\_bicyclic} $ \le 5.643 $
    \end{tabular}}     &    \multicolumn{1}{c|}{\begin{tabular}[c]{@{}c@{}} \texttt{NumAliphaticRings} $> 1.414 $, \\
     \texttt{SlogP\_VSA4} $> 2.059 $
    \end{tabular}}                 \\ 
    \cmidrule{2-4}
    & Support & \multicolumn{1}{c|}{158}     &      184                       \\ 
    \cmidrule{2-4}
    & Confidence       & \multicolumn{1}{c|}{0.898}    &      0.81                 \\ 
    \cmidrule{2-4}
    & Fitness & 0.247 & 0.223    \\
    \midrule
    \rowcolor[HTML]{E6E6E6}
    \multicolumn{4}{|c|}{\textbf{MNIST digit classification}}
    \\
    \midrule
    \multirow{4}{*}{2} & 
    \begin{tabular}[c]{@{}c@{}} Rules ($\mathbb{X}$) \\
    Target: $c = 7$  
    \end{tabular} & 
    \begin{tabular}[c]{@{}c@{}}  $z_{44} \ge 6.223$, \\
    $0 \le z_{114} \le 3.105$
    \end{tabular} &  
    \begin{tabular}[c]{@{}c@{}}   $z_{44} \ge 6.429$, \\  
    $z_{114} \le 3.463$
    \end{tabular}  \\
    \cmidrule{2-4}
    & Support & 5340 & 5518    \\
    \cmidrule{2-4}
    & Confidence & 0.900 & 0.891    \\
    \cmidrule{2-4}
    & Fitness & 0.682 & 0.687    \\
    \midrule
    \rowcolor[HTML]{E6E6E6}
    \multicolumn{4}{|c|}{\textbf{Brain tumour MRI classification}}
    \\
    \midrule
\multirow{4}{*}{\begin{tabular}[c]{@{}c@{}} $2$   \end{tabular} } & \begin{tabular}[c]{@{}c@{}} Rules ($\mathbb{X}$) \\
Target: $c = 2$ \end{tabular} & \begin{tabular}[c]{@{}c@{}}  $z_{1200} \ge 0.653$, \\ $-0.208 \le  z_{1110} \le -0.007$ \end{tabular} & \begin{tabular}[c]{@{}c@{}}   $z_{115} > 0.781$ \end{tabular} 
\\ 
\cmidrule{2-4} 
                       & Support                                                                                                                                                          & 301                                                                                                                                                              & 274                      \\ 
\cmidrule{2-4} 
                       & Confidence                                                                                                                                                       & 0.934                                                                                                                                                              & 0.934\\ 
\cmidrule{2-4} 
                       & Fitness                                                                                                                                                          & 0.853                                                                                                                                                                & 0.778    \\
    \bottomrule
    \end{tabular}
    \label{tab:global_results}
\end{table*}

\begin{figure*}
    \centering
    \includegraphics[width=\linewidth]{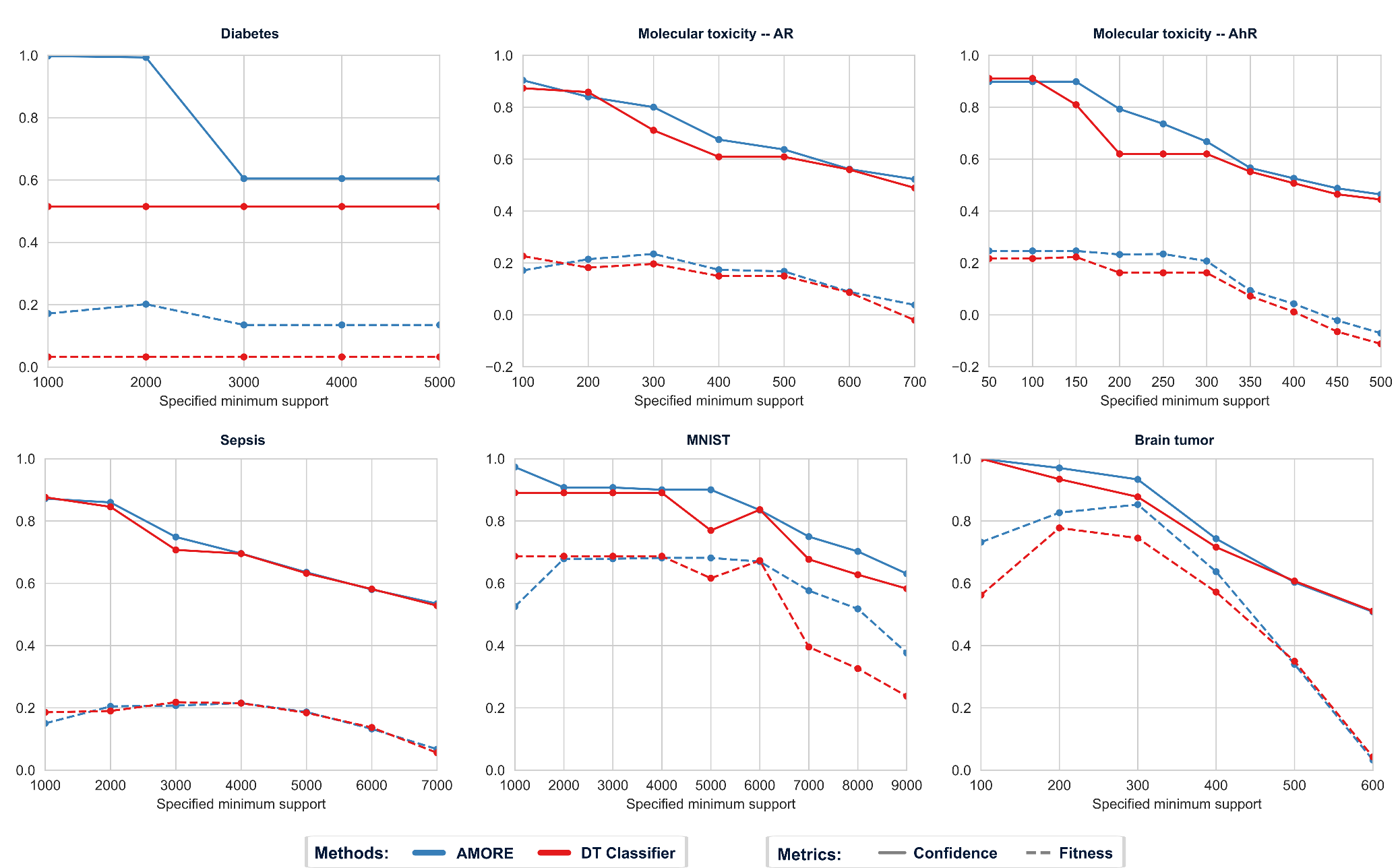}
    \caption{Comparing \ac{AMORE} and DT classifiers while specifying the same minimum support, $s_{min}$, on the $x$ axis and fixing the $l_{max}$ as in \cref{tab:global_results} for each task. 
    We choose the range of $s_{min}$ according to the size of target subgroup for each task. We perform a grid search for other hyperparameters for both methods in all cases and compare the confidence and fitness of their best rule sets. We observe that \ac{AMORE} generally yields better or equivalent results compared to the DT classifier.}
    \label{fig:compare_DT}
\end{figure*}


\subsection{Baseline method}

In all of our experiments, we chose the \ac{DT} classifier as the baseline method because, similar to our method, it is capable of model-agnostic rule extraction and does not require predefined discretization for numerical features. We were unable to find another method with such capabilities. Additionally, it allows us to verify the expected difference between the global and regional rule extraction. 

The \ac{DT} classifiers were trained to classify samples belonging to the target subgroup as positive class and others as negative class. The rules from a \ac{DT} classifier are formed by the decision path to the node with the highest fitness. The rules from \ac{AMORE} are the rule set with the highest fitness among all candidate rule sets. All results are provided in \Cref{tab:global_results}. 

We set the same random seed for training the classifiers of all tasks. Results from \ac{AMORE} are deterministic when both the model and training set remain unchanged. Similarly, the \ac{DT}  yields consistent results upon varying the random seeds, as all features are involved in split selection in decision trees. Therefore, there is no variance reported in \cref{tab:global_results}.  

\subsection{Hyperparameters}
We configure a same $l_{max}$ for both methods in each task. However, the hyperparameter $s_{min}$ is highly correlated to the optimal fitness that a rule set can achieve. Therefore, we conducted a grid search for  $s_{min}$, alongside other method-specific hyperparameters, for both methods in each task. We provide the selected hyperparameters for both methods in Table 2 in Supplementary Material. \Cref{fig:compare_DT} visualizes the comparison between the two methods while varying the value of $s_{min}$. 

To avoid selecting rules that have slightly higher fitness but significantly lower confidence, we specify a \emph{confidence lower bound} $\iota$ for choosing the best rule set during grid search in all experiments. This lower bound filters out any rule sets with confidence levels below it when selecting the best rule set based on fitness. However, if no rule set with a confidence above this lower bound is obtained, we simply select the one with the highest fitness. To verify the robustness of \ac{AMORE} with respect to  $\iota$, we conducted a sensitivity analysis by varying $\iota$ from 0.7 to 0.9 (Figure 1 in Supplementary Material). With different $\iota$, \ac{AMORE} consistently obtains better or equivalent performance compared to \ac{DT} classifiers. We chose $\iota = 0.8$ as it is considered reasonably good for rule confidence while allowing a higher support.

\setlength{\tabcolsep}{4pt}
\begin{table*}[!ht]
\centering
\footnotesize
\caption{Rules extracted by \ac{AMORE} for interpreting predictions of individual test samples with different levels of model uncertainty in the diabetes dataset. We set $s_{min} = 1000$, $l_{max}=3$ for all test samples. The prediction threshold $y_{th}$ is 0.457. The rules exhibit lower confidence and fitness when the predicted probability $\hat{y}_p$ is lower.}
\begin{tabular}{|c|c|c|c|}
\toprule
\rowcolor[HTML]{E6E6E6}
 & \textbf{Test sample 1} & \textbf{Test sample 2} & \textbf{Test sample 3} \\
 \toprule
Predicted probability & $\hat{y}_p = 0.908$ & $\hat{y}_p = 0.591$ & $\hat{y}_p = 0.424$ \\
\midrule
\begin{tabular}[c]{@{}c@{}} Rules ($\mathbb{X}$) \\
Target: $\hat{y} = c$ \end{tabular} & 
\begin{tabular}[c]{@{}c@{}} \texttt{HbA1c\_level} $ \ge 6.25$, \\ 
\texttt{blood\_glucose\_level} $\ge 190.$,\\
\texttt{age} $> 40$
\end{tabular} &  
\begin{tabular}[c]{@{}c@{}} \texttt{HbA1c\_level} $ \ge 5.071$,\\
\texttt{blood\_glucose\_level} $\ge 174.286$ \\
$33.384 \le \texttt{BMI} \le 45.071$  
\end{tabular} & 
\begin{tabular}[c]{@{}c@{}} 
\texttt{HbA1c\_level} $ \ge 5.253$, \\ 
\texttt{age} $ > 54$, \\
\texttt{BMI} $ \ge 37.48$
\end{tabular} \\
\midrule
Support & 1967 & 1128 & 1340    \\
\midrule
Confidence & 1. & 0.974  & 0.893  \\
\midrule
Fitness & 0.147  & 0.08     &  0.079   \\
\bottomrule
\end{tabular}
\label{tab:diabetes_local}
\end{table*}

\subsection{Task 1 -- Diabetes prediction}\label{sec:diabetes}

The Diabetes prediction dataset is a public dataset \cite{diabetes@kaggle} that
contains medical and demographic information of 100,000 patients along with their diabetes status. It consists of eight raw features including two binary features (\texttt{hypertension}, \texttt{heart disease}), two categorical features (\texttt{gender}, \texttt{smoking history}), and four numerical features (\texttt{age}, \texttt{BMI}, \texttt{HbA1c\_level}, and \texttt{blood\_glucose\_level}).

We trained a logistic regression model to classify whether a patient has diabetes or not, and then extracted rules that interpret the key knowledge acquired by the classifier in recognizing diabetes patients. We first identify highly influential features using our feature selection method (\Cref{alg:find_freq_features}) and then extract rules using those features.
The most informative feature identified by both methods is HbA1c level (\texttt{HbA1c\_level}). As a measure of a person's average blood sugar level over the past 2-3 months, it reasonably emerges as a key factor in predicting diabetes, demonstrating the effectiveness of our feature selection method. As only one rule is allowed in this task, \ac{AMORE} achieved much better performance than the decision tree classifier  (\Cref{tab:global_results}) due to the trade-off issue in global rule extraction is magnified in such scenarios.

Furthermore, we demonstrate obtaining local rules for three test samples by our method, as shown in \Cref{tab:diabetes_local}: one with a very high predicted probability, another close to the decision threshold, and a third below the threshold. 
The obtained local rules exhibit lower confidence and fitness when the predicted probability is lower, indicating that the quality of local rules correlates with  the model's uncertainty on the given sample. For samples that are distant from the target subgroup (e.g. $\hat{y}_p \approx 0$), our approach may not return valid local rules (rules that give a higher confidence than without them).

\subsection{Task 2 -- Sepsis prediction}

The Sepsis dataset \cite{reyna2019early} 
is a public dataset consisting of hourly vital signs and lab data, along with demographic data, for 40,336 patients obtained from two distinct U.S. hospital systems.  
The objective is to predict the onset of sepsis within 72 hours from the time a patient has been admitted to the ICU. 
We treat the data as time series, with each data sample comprising 72 time steps (hours), and each time step including 34 features representing vital signs and laboratory data. For patients who do not have records for the full 72 hours, we treated features of missing hours as missing values filled by NaN before feature augmentation. 

We conducted feature augmentation by three cumulative operations on the 34 original features: i). \texttt{feature\_ct} denotes the cumulative number of time steps that a raw feature has non-missing values; ii). \texttt{feature\_cmax} denotes the maximum value of a specific feature among all passed time steps; iii). \texttt{feature\_csum} denotes the value sum of a specific feature over passed time steps. We added the time index as a suffix to the augmented feature. For example, \texttt{HR\_ct\_t45} denotes the cumulative number of time steps that have recorded \texttt{Heart Rate} (beats per minute) at the $45$-th time step. 
After augmentation, each time step has 136 features, resulting in a total of 9,792 features for each sample. 

We trained a \acf{NCDE} model \cite{kidger2020neural} on this dataset to predict the onset of sepsis. \ac{NCDE} is able to deal with irregular sampled time series using controlled differential equations. \ac{ODE} solvers are involved in the approximation of the gradients of the model parameters. This way, we demonstrate that our approach can also be applied to this type of models. More information about \ac{NCDE} can be found in Supplementary Materials.

As this task has a high dimensional feature space, our method selected different features with decision trees. We also observe that the top rules of both methods are from features augmented by cumulative time steps. This suggests that the model considers the frequency of taking those vital signs and lab tests as key information for predicting sepsis. The extracted rules can be found in \Cref{tab:global_results}. The feature \texttt{Temp} indicates temperature ($^\circ\mathrm{C}$) and \texttt{SPB} is Systolic BP (mm Hg).

\begin{figure*}[!t]
    \centering   
    \includegraphics[width=0.8\textwidth,trim=0.cm 0.cm 0.cm 0.cm,clip]{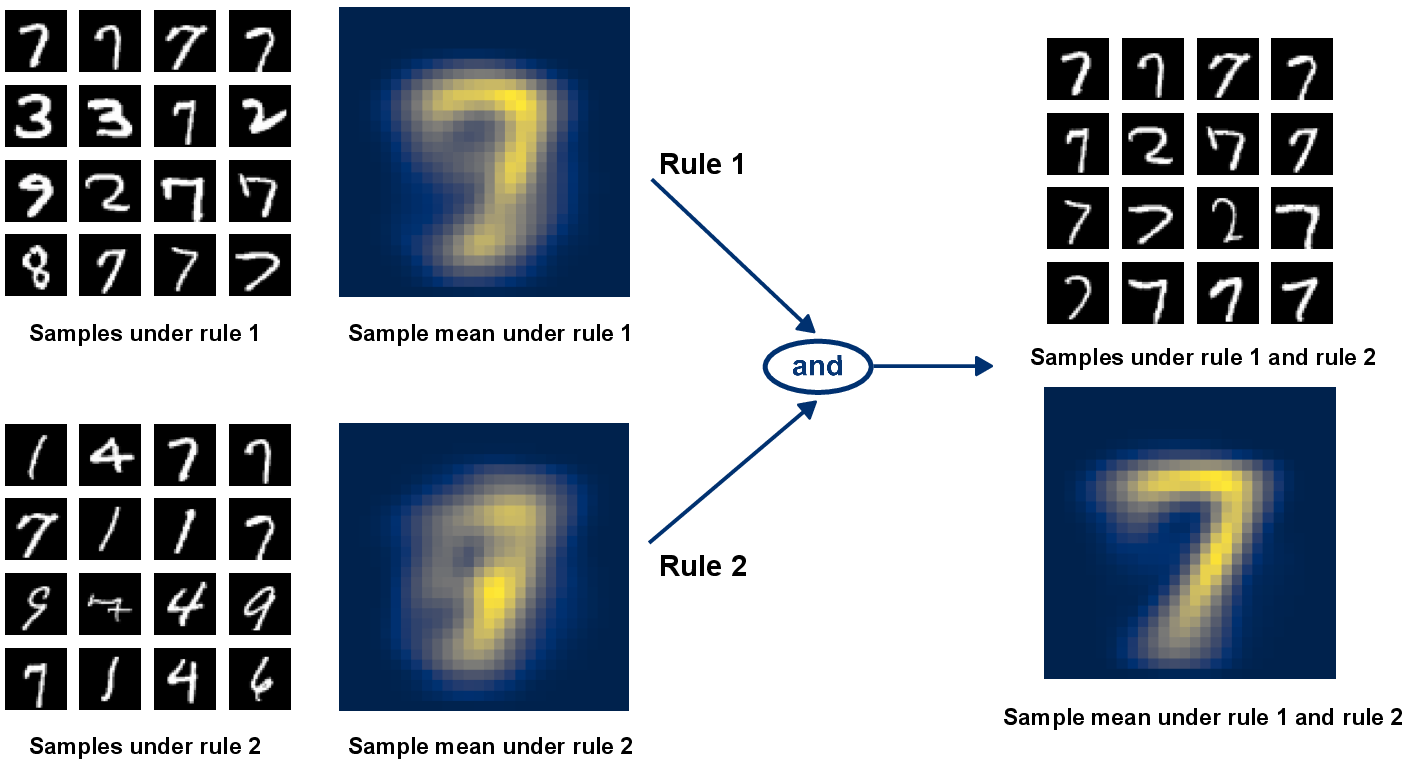}  
    \caption{Visualization of extracted rules from the MNIST dataset. Here, rule 1 is $z_{44} \ge 6.223$ and rule 2 is $0 \le z_{114} \le 3.105$ as in \Cref{tab:global_results}.}
    \label{fig:mnist}
\end{figure*}

\begin{figure*}[!h]
    \centering    \includegraphics[width=0.8\textwidth,trim=0.cm 0.cm 0.cm 0.cm,clip]{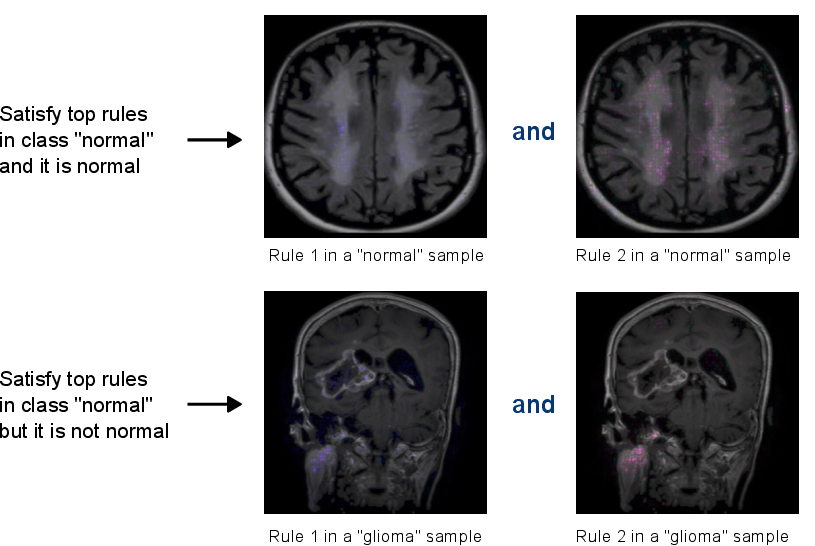}  
    \caption{Visualization of extracted rules from brain tumor MRI dataset by two samples that satisfy the extracted rules of class ``normal". The above row is a sample truly ``normal" and the below row is a sample not labeled as ``normal". The first column highlights pixels that have higher contributions to the latent state ($z_{1200}$) in rule 1 ($z_{1200} \ge 0.653$) and the second row is for the latent state ($z_{1110}$) in rule 2 ($-0.208 \le z_{1110} \le -0.007$). We visualize the latent state of each rule by coloring pixels according to their impact scores for the latent state, highlighting highly impactful regions of each rule in the figures.}
    \label{fig:braintumour_samples}
\end{figure*}

\subsection{Task 3 -- Molecular toxicity prediction}
Predicting the toxicity of new medicines using molecular structure is crucial in drug discovery, given that over 30\% of promising pharmaceuticals fail in human clinical trials due to toxicity discovered in animal models \cite{kola2004can}. The Tox21 (The Toxicology in the 21st Century) challenge contained in the MoleculeNet \cite{wu2018moleculenet} 
is a public dataset that assesses the toxicity of 8,014 compounds across 12 different categories. These categories include 7 nuclear receptors and 5 stress response pathways. 

We trained a graph attention network \cite{velivckovic2017graph} to predict the categories of compounds using their chemical structures as input. This is a multi-label classification task, and we chose the first two categories for rule extraction as demonstration: the androgen receptor (AR) and aryl hydrocarbon receptor (AhR). 
We utilize RDKit descriptors \cite{rdkit2023packages} to map molecule structures to interpretable features, providing a consistent set of features that describe essential molecule properties. 
As the results presented in \Cref{tab:global_results},
\ac{AMORE} obtained higher fitness than DT classifiers in both cases.  

\subsection{Task 4 --MNIST digits classification}

In addition to tabular data, we showcase the capability of our method to interpret models for image data by using MNIST dataset \cite{lecun2010mnist}. 
We trained a neural network with a convolutional layer and two fully connected layers for this task. The target class is set to $c=7$, i.e., we identify the model's key knowledge for classifying the digit 7. We used the model's 
latent representations (input to the final linear layer) as features to generate rules, which dimension is 128.
The performance of extracted rules can be found in \Cref{tab:global_results} as well. For a visual demonstration, we illustrate the extracted rules by samples that satisfy each rule in \Cref{fig:mnist}. It is clear to see that the first rule emphasises the upper half of digit 7, while the second rule focuses on the lower half.

\subsection{Task 5 -- Brain tumour MRI classification}

The brain tumour MRI dataset \cite{chitnis2022brain} is a public dataset that includes brain MRI images with four classes: \emph{glioma tumour, meningioma tumour, pituitary tumour, and normal}. We used an EfficientNet \cite{tan2019efficientnet} architecture for this classification task which is pre-trained on ImageNet-1K dataset \cite{imagenet15russakovsky}. 

Like the MNIST task, we also use the latent representations from the model as features to generate rules in this experiment, which dimension is 1280. We extracted rules for identifying ``normal" brain MRI images where the class ``normal" is the minority class in this dataset. The sample means of the MRI images are not visually interpretable. Therefore, we showcase two samples that satisfy rules extracted by \ac{AMORE} in \Cref{fig:braintumour_samples}: one is from class ``normal" and the other is malignant. We visualise each rule by colouring pixels according to each pixel's impact on the latent feature that defines the rule, thereby highlighting the most impactful areas for each rule in an MRI image.

\section{Discussion}

In the presented experiments, we illustrate that \ac{AMORE} can extract representative rules with large support and high confidence for specific data subgroups, achieving better or equivalent performance compared to DT based explanation. We showcase the best conjunctive rule sets (i.e. with only ``\emph{AND}" operations) here, a more comprehensive view could be constructed by post-processing, for example, using a sequential covering strategy or merging candidate rule sets by ``\emph{OR}" operations. 

While two hyperparameters are configured to limit the minimum support ($s_{min}$) and the maximum length ($l_{max}$) of extracted rules for both methods, we observe that the rules extracted by \ac{AMORE} often have an actual support and length that are closer to these configurations. This aligns with our expectations to have more precise control of extracted rules in regional rule extraction.

We demonstrated how to apply \ac{AMORE} with data types other than tabular data in several tasks. 
For instance, in the molecular toxicity prediction task, we map the molecular structures to RDKit descriptors, providing interpretable features by the descriptors. Likewise, in the MNIST and Brain tumor experiments, we extract rules from the latent space of models and then interpret them through visualization in the raw pixel space. In general, for data with features that are not directly comprehensible in the form of logical rules, one can map the raw features to an interpretable feature space so that \ac{AMORE} can be applicable. For instance, the speech and text data can be transformed by mapping raw features to semantic concepts for interpreting emotion classifiers. However, it is challenging to make such a transformation for time series if we expect interpretable features that can describe complex longitudinal properties. We consider this as future work.

Regarding regression tasks and unsupervised tasks with continuous target variables, a target subgroup can be defined by a value range of its target variable, such as a quantile of the variable. However, \ac{AMORE} is not intended to reveal the dynamic patterns of a continuous variable, such as patterns indicating an increase or decrease of the variable. Future work may consider the use of the dynamics of the continuous variable (e.g., the first or second-order derivatives) to enable extracting rules for dynamic patterns.




\bibliographystyle{IEEEtran}
\bibliography{references.bib}




\clearpage

\section{Supplementary}  

\setcounter{figure}{0}
\setcounter{table}{0}

\setlength{\tabcolsep}{3pt}

\setlength{\tabcolsep}{3pt}
\begin{table*}[!h]
\centering
\footnotesize
\caption{Summary of all tasks in our experiments. The training size is the number of samples in the training set. We use 70\% of each dataset as the training set and the remaining 30\% data were used for validation and testing. We did not conduct cross-validation to select hyperprameters for these predictive models as they are not our focus in this work and our method is model-agnostic. ``\#Features" is the number of features, accounting for features involved in rule extraction, so it may not be the number of raw features. For the Diabetes and Sepsis tasks, it is the number of features after preprocessing; for Molecular tasks, it is the number of explainable features mapped from molecular graphs; for the MNIST and brain tumor tasks, it corresponds to the number of latent states we used for generating rules. The target class $\bm{c}$ represents the class of interest. $\bm{\#(y^*=c)}$ and $\bm{\#(\hat{y} = c)}$ denote the number of samples with true labels as $\bm{c}$ and those predicted as $\bm{c}$, respectively. Recall, precision, and AUC are given by the predictive model over the training set as we extract rules from the training set.  $\text{AUC}_{test}$ is obtained over the test set.}
\begin{tabular}{|c|c|c|c|c|c|c|c|c|c|c|}
\toprule
\rowcolor[HTML]{D6D6D6}
\textbf{Task} & \textbf{Model} & \textbf{Training size} & \textbf{\#Features} & \ \ $\bm{c}$ \ \ & $\bm{\#(y^*=c)}$ & $\bm{\#(\hat{y} = c)}$ & \textbf{Recall} & \textbf{Precision} & \textbf{AUC} & $\textbf{AUC}_{test}$ \\
\midrule
Diabetes & Logistic regression & 70000   & 15  & 1  & 5953  & 13379 & 0.897 & 0.399 & 0.962 & 0.962 \\
\midrule
Sepsis & \acs{NCDE} & 32266   & 9792  & 1  &  1741  &  6299 & 0.766 & 0.212 & 0.876 & 0.844 \\
\midrule
{\begin{tabular}[c]{@{}c@{}} Molecular\\ AR \end{tabular}} & GNN  & 5481   & 192  &  1 &  219  & 834  & 0.731 & 0.192 & 0.887 & 0.579\\
\midrule
{\begin{tabular}[c]{@{}c@{}} Molecular\\ AhR \end{tabular}} & GNN  &  5481  & 192  &  1 &  162  & 511  & 0.846  & 0.268 & 0.948 & 0.877\\
\midrule
MNIST & CNN & 60000   & 128  & 7  & 6265   &  6274 & 0.997 & 0.996 & 1 &  1\\
\midrule
{\begin{tabular}[c]{@{}c@{}} Brain tumor \\ MRI \end{tabular}} & EfficientNet & 2160   & 1280  & 2  & 306   &  306 & 1 & 1 & 1 &  0.999\\
\bottomrule
\end{tabular}
\label{tab:task_summary}
\end{table*}

\begin{table*}[!h]
\centering
\caption{Hyper-parameters configured for \ac{AMORE} and {DT} classifier in all experiments. We picked several key hyper-parameters of DT trees to tune and all other parameters are the default values set in scikit-learn package. All the hyperparameters are selected by grid search.}
\begin{tabular}{|c|cccc|cccc|}
\toprule
\multirow{2}{*}{Task} &  \multicolumn{4}{c|}{\cellcolor[HTML]{D6D6D6}AMORE}         & \multicolumn{4}{c|}{\cellcolor[HTML]{D6D6D6}DT  Classifier}                               \\ 
 \cmidrule{2-9} 
 & \cellcolor[HTML]{E6E6E6}$l_{max}$ & \cellcolor[HTML]{E6E6E6}$s_{min}$ & \cellcolor[HTML]{E6E6E6}$n_g$ & \cellcolor[HTML]{E6E6E6} bin\_strategy & \cellcolor[HTML]{E6E6E6} max\_depth & \cellcolor[HTML]{E6E6E6} \cellcolor[HTML]{E6E6E6} min\_samples\_leaf & \cellcolor[HTML]{E6E6E6} class\_weight & \cellcolor[HTML]{E6E6E6} criterion \\ \hline
Diabetes              & 1         & 2000    & 7    & uniform   & 1          & 1000               & uniform                   & gini      \\ \midrule
Sepsis                & 2         & 2000    & 7     & uniform   & 2          & 2000               & uniform                   & gini      \\ \midrule
Toxicity -- AR        & 2         & 300     & 3    & kmeans   & 2          & 100                & uniform                   & gini      \\ \midrule
Toxicity -- AhR    & 2         & 50    & 3    & kmeans  & 2          & 150               & balanced                   & gini      \\ \midrule
MNIST                 & 2         & 4000    & 6    & kmeans   & 2          & 1000               & uniform                   & entropy      \\ \midrule
Brain tumor MRI                & 2         & 300    & 8    & kmeans   & 2          & 200               & uniform                   & gini      \\ 
\bottomrule
\end{tabular}
\label{tab:hyperparam}
\end{table*}

\subsection{Summary of all tasks}
In this section, we provide details of the predictive models applied in each task (\cref{tab:task_summary}) and the selected hyperparameters of both rule extraction methods (\cref{tab:hyperparam}).

\subsection{Sensitivity analysis of confidence lower bound}

In order to verify the robustness of \ac{AMORE} with respect to the confidence lower bound $\iota$, we conduct a sensitivity analysis by varying $\iota$ from 0.7 to 0.9.
We compare \ac{AMORE} and DT classifiers while varying the minimum support under different confidence lower bounds. We can see that changing $\iota$ does not affect the diabetes and brain tumor tasks. For other tasks, the differences are not significant and \ac{AMORE} still demonstrates better or equivalent performance compared to DT classifiers across different $\iota$. These results are shown in \Cref{fig:sensitivity_analysis}.

\begin{figure*}[!t]
    \centering   
    \subfloat[Diabetes prediction]
    {\includegraphics[width=0.85\textwidth,trim=0.cm 0.cm 0.cm 0.cm,clip]{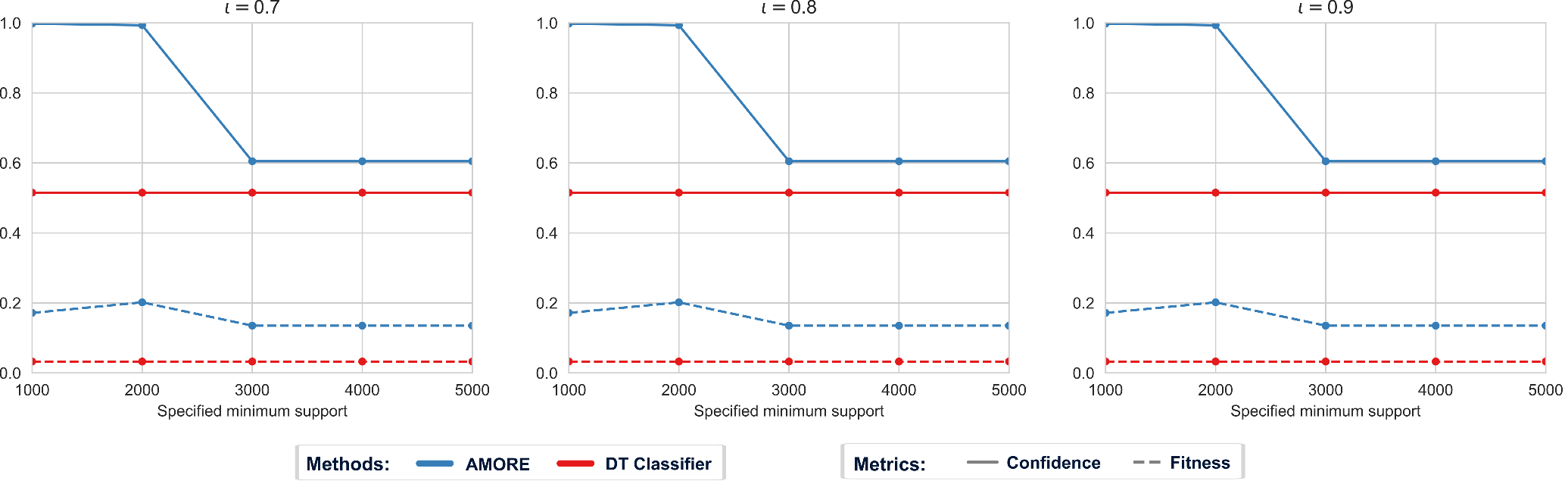}  
    \label{fig:compare_DT_diabetes}}
    \\
    \subfloat[Sepsis prediction]
    {\includegraphics[width=0.85\textwidth,trim=0.cm 0.cm 0.cm 0.cm,clip]{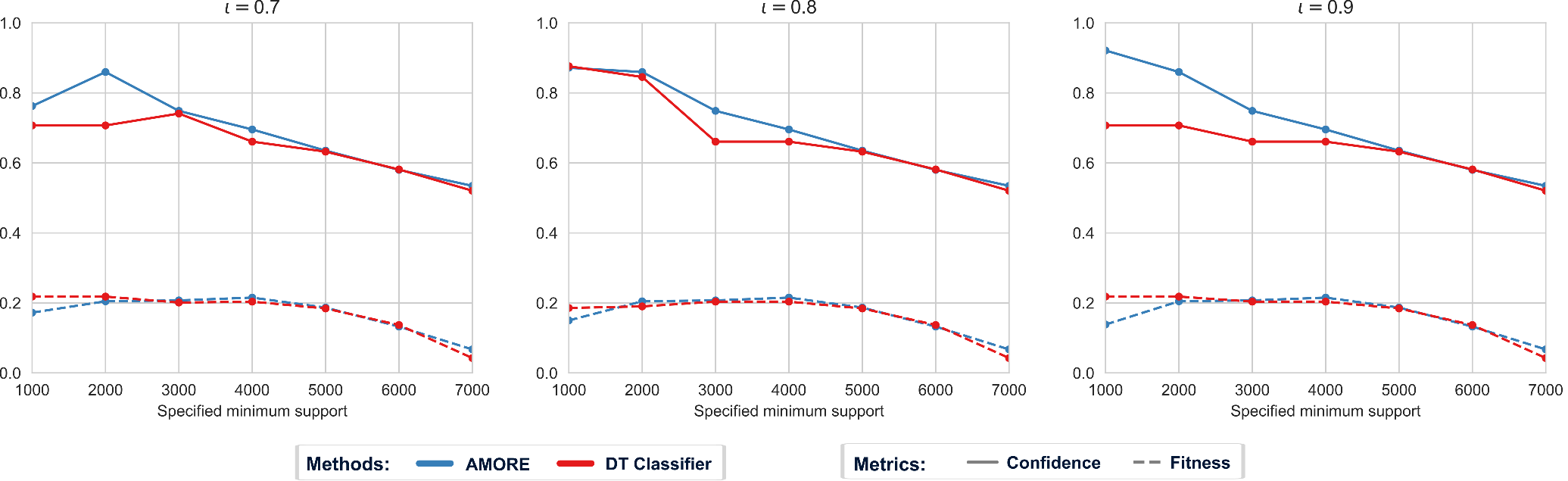} 
    \label{fig:compare_DT_sepsis}}
    \\
    \subfloat[Molecular toxicity prediction -- AR]
    {\includegraphics[width=0.85\textwidth,trim=0.cm 0.cm 0.cm 0.cm,clip]{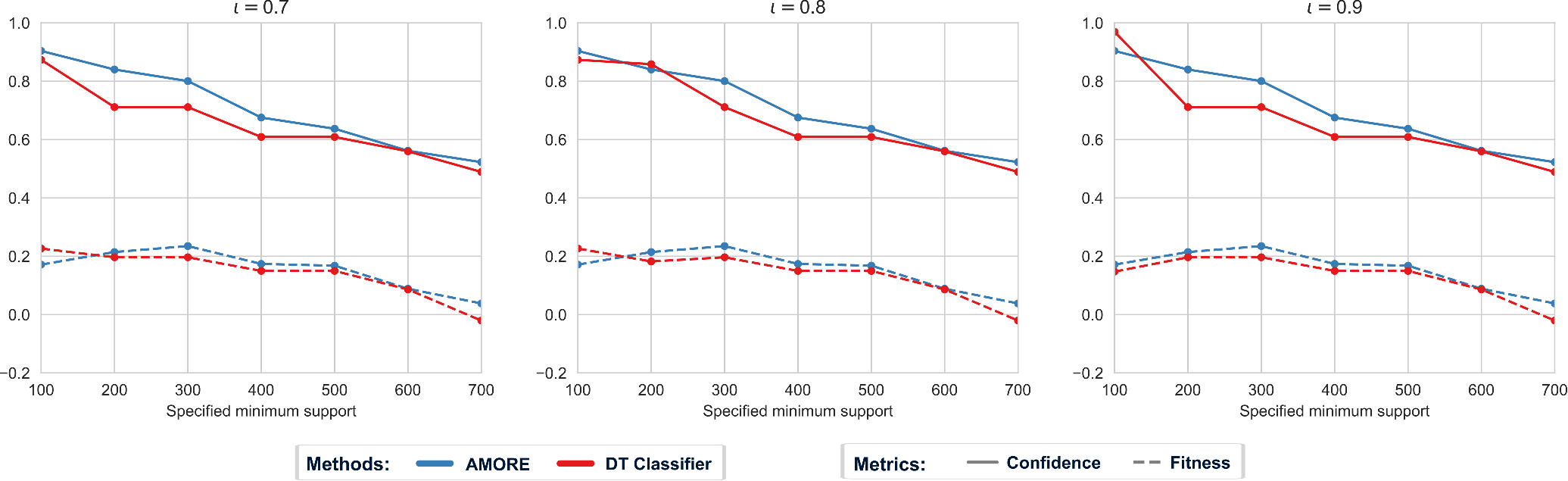} 
    \label{fig:compare_DT_AR}}
    \\
    \subfloat[Molecular toxicity prediction -- AhR]
    {\includegraphics[width=0.85\textwidth,trim=0.cm 0.cm 0.cm 0.cm,clip]{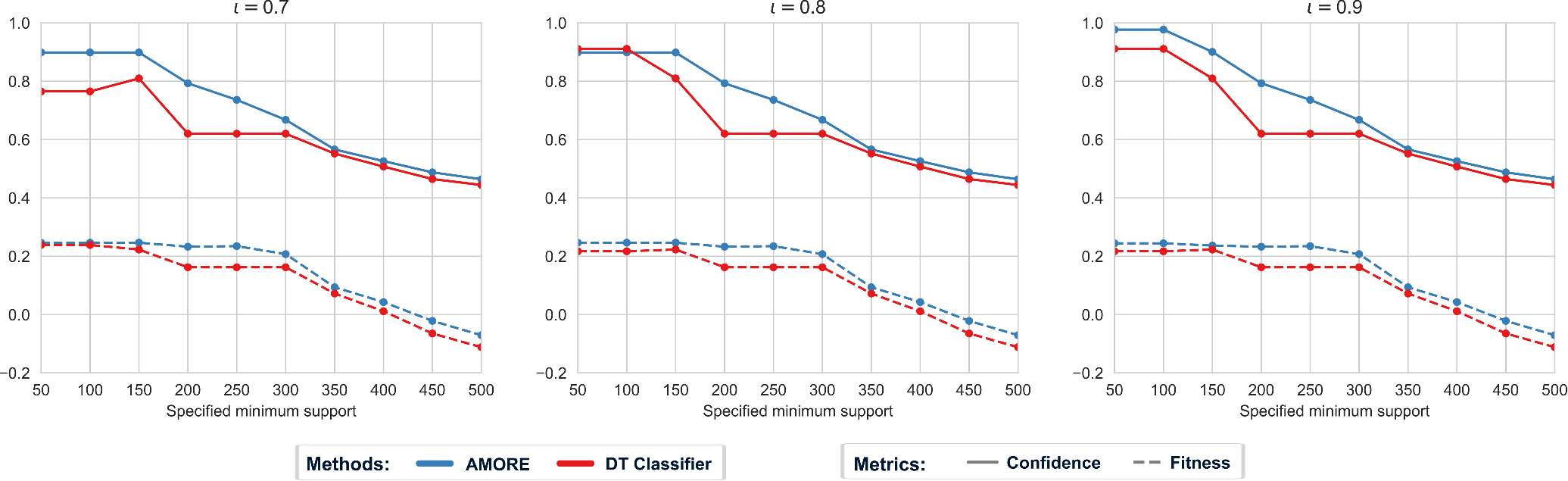} 
    \label{fig:compare_DT_AhR}}
\caption{The Sensitivity analysis of confidence lower bound $\iota$ (to be continued). We compare \ac{AMORE} and DT classifiers while varying the minimum support under different confidence lower bounds. The confidence lower bound $\iota$ is set to [0.7, 0.8, 0.9] for all tasks. We can see that changing $\iota$ does not affect the diabetes and brain tumor tasks. For other tasks, the differences are not significant and \ac{AMORE} still demonstrates better or equivalent performance compared to DT classifiers.}
\label{fig:sensitivity_analysis}
\end{figure*}

\setcounter{subfigure}{4}
\begin{figure*}[!t]
    \ContinuedFloat
    \centering   
    \subfloat[MNIST -- digit 7]
    {\includegraphics[width=0.9\textwidth,trim=0.cm 0.cm 0.cm 0.cm,clip]{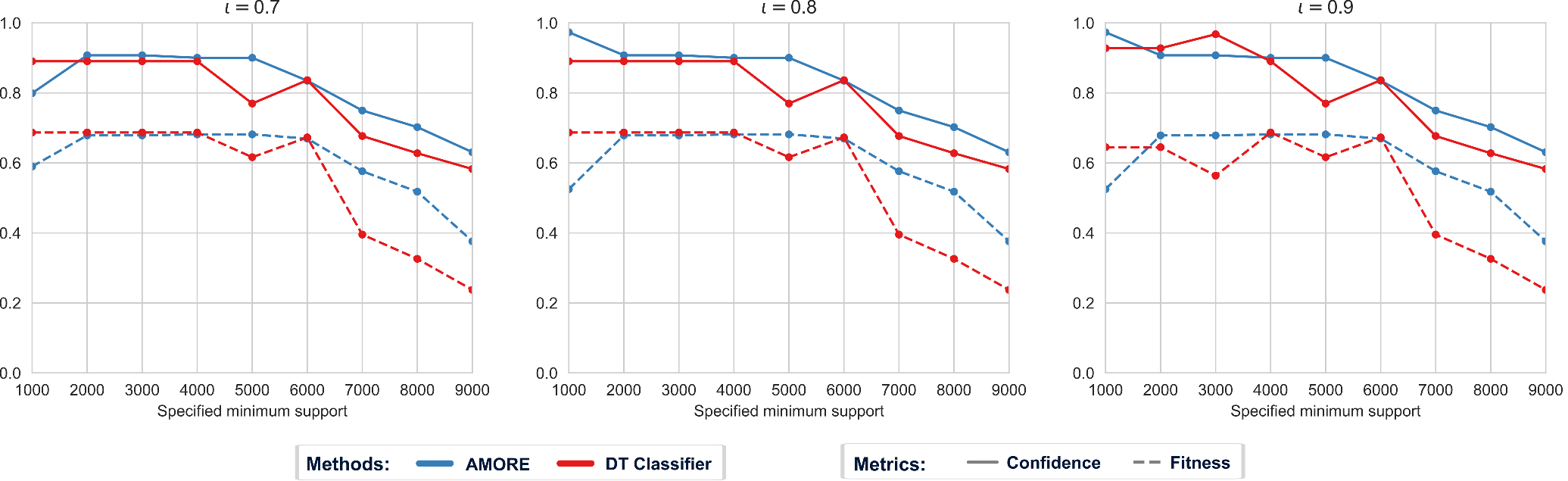} 
    \label{fig:compare_DT_mnist}}
    \\
    \subfloat[Brain tumor MRI -- normal]
    {\includegraphics[width=0.9\textwidth,trim=0.cm 0.cm 0.cm 0.cm,clip]{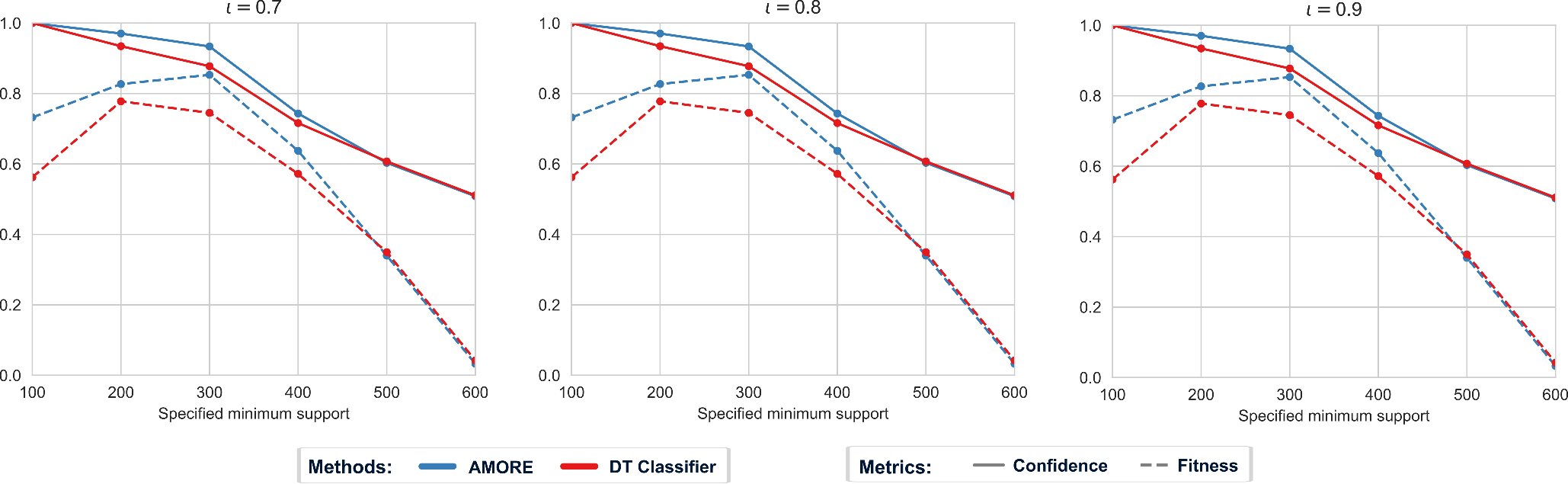} 
    \label{fig:compare_DT_braintumor}}
\caption{The Sensitivity analysis of confidence lower bound $\iota$. (Continued)}
\end{figure*}

\subsection{\acf{NCDE}}\label{sec:ncde}
\ac{NCDE} is a type of models that use neural networks to simulate the dynamic function of variables and use ODE solvers to estimate integrals for the dynamic function. It is based on \ac{NODE} \cite{chen2018neural} which can use neural networks to approximate a latent state that has no explicit formulation with its dynamics. For instance: $ z(t_1) = z(t_0) + \int_{t_0}^{t_1} f_{\theta}(z(t), t) dt$.
Here $z(t)$ denotes latent states at time $t$, $f_{\theta}(\cdot)$ represents an arbitrary neural network.
\ac{NCDE} \cite{kidger2020neural} uses a controlled different equation, in which the dynamics of $z(t)$ are assumed to be controlled or driven by an input signal $x(t)$. 
\begin{equation}
\begin{split}
    z(t_1) &= z(t_0) + \int_{t_0}^{t_1} \tilde{f}_{\theta}(z(t), t) dx(t) \\
    &= z(t_0) + \int_{t_0}^{t_1} \tilde{f}_{\theta}(z(t), t) \frac{dx}{dt} dt, \\
\end{split}
\end{equation}
where $\tilde{f}_{\theta}: R^{H} \rightarrow R^{H\times D}$, $H$ is the dimension of $z$, $D$ is the dimension of $x$. By treating $f_{\theta}(z(t), t) \frac{dx}{dt}$ as one function $g_{\theta}(z(t), t, x)$, one can solve \ac{NCDE} by regular methods for solving \ac{NODE}. 
The trajectory of control signals ($x(t)$) in \ac{NCDE} can be fitted independently with  $\tilde{f}_{\theta}$, which enables \ac{NCDE} to deal with irregular sampled time series and work with online streaming data. 

For this type of models, we can obtain the Jacobian ${\partial z}/{\partial x}$ easily when we i) use observations as control signals; ii) $x(t)$ is approximated by an invertible function. According to the chain rule:
\begin{equation}
    \begin{split}
        \frac{\partial z}{\partial x} = \frac{\partial z}{\partial t} \frac{dt}{dx} = \left( \tilde{f}_{\theta}(z(t), t) \frac{dx}{dt} \right)\frac{dt}{dx} = \tilde{f}_{\theta}(z(t), t)
    \end{split}
\end{equation}
This way only needs a forward computation instead of a back-propagation to obtain the Jacobian ${\partial z}/{\partial x}$.

\end{document}